\documentclass[10pt]{article} 
\usepackage{amsmath, amssymb, amsthm}

\usepackage{algorithm}
\usepackage{algorithmic}
\usepackage{url}

\usepackage{newfloat}
\usepackage{listings}
\floatstyle{ruled}
\newfloat{listing}{tb}{lst}{}
\floatname{listing}{Listing}

\usepackage{subfigure}
\usepackage{multirow}

\usepackage{microtype}
\usepackage{graphicx}
\usepackage{booktabs} 

\usepackage{hyperref}

\usepackage{xspace}
\usepackage{setspace}

\usepackage{pgfplotstable}
\usepackage{tikz}
\pgfplotsset{compat=newest}
\usetikzlibrary{calc,positioning,shapes.geometric,math}

\usepackage{etoolbox}
\newtoggle{release}
\togglefalse{release}
\iftoggle{release}{
\newcommand{\alex}[1]{}
\newcommand{\gab}[1]{}
\newcommand{\adios}[1]{}
}
{
\newcommand{\alex}[1]{{\color{blue} A: #1}}
\newcommand{\gab}[1]{{\color{green} G: #1}}
\newcommand{\adios}[1]{{\color{magenta}Y: #1}}
}

\newcommand{\diffq}{\textsc{DiffQ}\xspace}

\newcommand{\Q}{\mathbf{Q}}
\newcommand{\tQ}{\mathbf{\tilde{Q}}}
\newcommand{\N}{\mathbf{N}}
\newcommand{\M}{\mathbf{M}}

\newcommand{\real}{\mathbb{R}}
\newcommand{\nat}{\mathbb{N}}

\newcommand{\esp}[1]{\mathbb{E}\left[#1\right]}

\usepackage{amsmath,amsfonts,bm}

\def\eqref#1{equation~\ref{#1}}

\def\1{\bm{1}}

\DeclareMathAlphabet{\mathsfit}{\encodingdefault}{\sfdefault}{m}{sl}
\SetMathAlphabet{\mathsfit}{bold}{\encodingdefault}{\sfdefault}{bx}{n}
\newcommand{\tens}[1]{\bm{\mathsfit{#1}}}

\def\tQ{{\tens{Q}}}

\usepackage[accepted]{tmlr}

\title{Differentiable Model Compression via Pseudo Quantization Noise}

\author{\name Alexandre D\'efossez\thanks{Equal contribution.} \email defossez@fb.com \\
      \addr Meta AI, FAIR Team, Paris, France
      \AND
      \name Yossi Adi$^{*}$ \email adiyoss@fb.com \\
      \addr Meta AI, FAIR Team, Tel-Aviv, Israel
      \AND
      \name Gabriel Synnaeve \email gab@fb.com\\
      \addr Meta AI, FAIR Team, Paris, France}

\def\month{09}  
\def\year{2022} 
\def\openreview{\url{https://openreview.net/forum?id=DijnKziche}} 

\begin{document}

\maketitle

\begin{abstract}
We propose \textsc{DiffQ} a differentiable method for model compression for quantizing model parameters without gradient approximations (e.g., \emph{Straight Through Estimator}). We suggest adding independent pseudo quantization noise to model parameters during training to approximate the effect of a quantization operator. \textsc{DiffQ} is differentiable both with respect to the unquantized weights and the number of bits used. Given a single hyper-parameter balancing between the quantized model size and accuracy, \diffq optimizes the number of bits used per individual weight or groups of weights, in end-to-end training. We experimentally verify that our method is competitive with STE based quantization techniques on several benchmarks and architectures for image classification, language modeling, and audio source separation. For instance, on the ImageNet dataset, \diffq compresses a 12 layers transformer-based model by more than a factor of 8, (lower than 4 bits precision per weight on average), with a loss of 0.3$\%$ in model accuracy. Code is available at \href{https://github.com/facebookresearch/diffq}{github.com/facebookresearch/diffq}.
\end{abstract}

\section{Introduction}
\label{intro}

An important factor in the adoption of a deep learning model for real-world applications is how easily it can be pushed
to remote devices. It has been observed that larger models usually lead to better performance, for instance
with larger ResNets~\citep{he2016deep} achieving higher accuracies than smaller ones.
In response, the community has worked toward smaller, and more efficient models~\citep{tan2019efficientnet}. Yet
an EfficientNet-B3 is still almost 50MB, a considerable amount if the model is to be included in online applications,
or updated with limited network capabilities. For other applications, such as language modeling~\citep{vaswani2017attention} or source separation~\citep{defossez2019music}, the typical model size is closer to 1GB, ruling out any kind of mobile usage. Efficient model compression is thus important
for on device adoption of deep learning models.
Thus, we focus in the present work on reducing model size, rather than achieving computational gains.

The simplest method to reduce model size consists in decreasing the number of bits used to encode individual weights. For instance, using 16 bits floating point numbers halves the model size, while retaining a sufficient approximation of the set of real numbers, $\real$, to train with first-order optimization methods~\citep{micikevicius2017mixed}.
When considering lower precision, for instance, 8 or 4 bits, the set of possible values is no longer a good approximation of $\real$, hence preventing the use of first-order optimization methods. Specifically, uniform quantization requires using the \texttt{round} function, which has zero gradients wherever it is differentiable.

\looseness=-1
Quantization can be done as a post-processing step to regular training. However, errors accumulate in a multiplicative fashion across layers,  with a possibly uncontrolled decrease in the model accuracy.
\citet{courbariaux2016binarized} and later \citet{krishnamoorthi2018quantizing} propose to use a gradient Straight-Through-Estimator (STE)~\citep{bengio2013estimating} in order to provide a non-zero gradient to the original weights. This allows the model to adapt to quantization during training and reduces the final degradation of performance. However, \citet{fan2020training} noticed instability and bias in the learned weights, as STE is not the true gradient to the function.

The nature of quantization noise has been extensively studied as part of Analog-to-Digital Converters (ADC). In particular, a useful assumption to facilitate the design of post-processing filters for ADC is
the independence of the input value and the ``Pseudo Quantization Noise'' (PQN), as formalized by \citet{widrow1996statistical}. In this work, we show that it also applies to deep learning model quantization, and provides a simple framework in which the output and the quantized model size are both differentiable, without any use of STE. This allows to optimally set the number of bits used per individual weight (or group of weights) to achieve a trade-off between size and accuracy, in a single training and at almost no extra cost. Even when the number of bits to use is fixed, we show that unlike STE, using independent pseudo quantization noise does not introduce bias in the gradient and achieves higher performance.
Although PQN has been proposed before for quantization~\citep{baskin2018nice,baskin2018uniq}, it has never been used on its own without any need for STE or other quantization methods, while achieving state-of-the-art performance.

\paragraph{Our Contribution:} (i) With \diffq, we propose to use pseudo quantization noise {\bf only} to approximate quantization at train time, as a differentiable alternative to STE, both with respect to the unquantized weights and number of bits used.
\\(ii) We provide a differentiable model size estimate, so that given a single penalty level $\lambda$, \diffq optimizes the number of bits per weight or group of weights to achieve a given trade-off between model size and accuracy. \\ (iii) We provide extensive experimental validation using various models (ConvNets and Transformers) and domains (image classification, language modeling, audio source separation). We demonstrate the efficiency of \diffq both in providing small footprint models with comparable performance to the uncompressed ones, together with easy and stable optimization, using only one sensitive hyper-parameter. 
\vspace{-0.1cm}
\section{Related Work}
\vspace{-0.1cm}
\label{relatedwork}

Early network quantization methods focused on low-bitwidth networks such as BinaryNet~\cite{courbariaux2015binaryconnect, courbariaux2016binarized}, XOR-Nets~\cite{rastegari2016xnor}, or Ternary networks~\cite{li2016ternary, wu2018training}. Although these methods produce highly quantized models, their performance is not on par with uncompressed ones. To improve accuracies, higher bitwidth quantization methods were studied~\cite{jung2019learning, zhang2018lq, mishra2017wrpn}. These methods followed the STE approach~\cite{bengio2013estimating}. STE allows the gradients to be backpropagated through the quantizers and, thus, the network weights can be adapted with gradient descent~\cite{courbariaux2016binarized}.

Variational approaches were used to make the categorical distribution over quantized weights differentiable. \citet{louizos2018relaxed} uses
a Gumbel-softmax~\citep{jang2016categorical} but requires 2 hyper-parameters and has no bitwidth tuning. \diffq has a single hyper-parameter and supports automatic bitwidth tuning.
\citet{shayer2017learning} relies on a Central Limit Theorem (CLT) application, however this prevents
weights from converging to a deterministic value, which would break the assumptions of the CLT. With \diffq, weights are free to converge
to any optimal value. Finally \citet{ullrich2017soft} uses a gaussian mixture model
trained on top of the weights, adding significant complexity both in terms of code, and computation. In contrast, \diffq adds only
one penalty term to the loss, optimized along the rest of the model in an end-to-end fashion.

An alternative is to use a smoothed version of the quantization operator, possibly with a trained meta-network \citep{chen2019metaquant}, however as the smoothed operator converges to the true one, gradients will eventually be zero almost everywhere. \citet{gong2019differentiable} use a meta-network to provide gradients despite quantization. However, their implementation for training the meta-network still relies on STE.

Additive noise injection has been studied by \citet{baskin2018nice}, although only during the first few epochs, after which STE based approximation is used. This work was extented to non uniform quantization~\citep{baskin2018uniq}. In contrast, \diffq uses only noise injection, and as demonstrated in Results Section, achieves a better accuracy for an equivalent compression level than both methods. Non uniform quantization was also studied by~\citet{polino2018model}, but without differentiability with respect to the weights, with worse performance than \diffq. Additive noise was also studied in the context of image compression~\citep{balle2017end,choi2019variable} in order to provide a differentiable pseudo-quantization operator. However, those work rely on an explicit estimation of the quantized values entropy, in particular with respect to a distribution of images. This formalism breaks down when having to quantize a single model, not a distribution, and \diffq uses a simpler approach where the bitwidth is directly tuned. More recently, \citet{park2022nipq} extended our method for activation quantization.

\looseness=-1
An important contribution from \diffq is the automatic tuning of the bitwidth using mixed-precision. Other mixed-precision quantization methods are based on Reinforcement Learning \citep{wang2019haq, elthakeb2020releq, liu2021sharpness}, second-order optimization \citep{dong2019hawq, dong2020hawq, yao2021hawq}, and differentiable quantization methods \citep{uhlich2020mixed, wang2020differentiable}. Comparing to \diffq, such methods are more complex (e.g., require plenty of parameter tuning), more computationally heavy, and most importantly based on STE approximations. \citet{wang2019haq, elthakeb2019releq} suggested learning a bitwidth assignment policy using reinforcement learning methods. In contrast, our method select bitwidth along training, using only first order optimization. \citet{jain2019trained, esser2020learned}, and \citet{bhalgat2020lsq} proposed learning the quantizer step-size or dynamic-range using STE, but do not allow to select the bitdwidth. Our experiments show that \diffq outperforms \citep{esser2020learned} (LSQ) both on most vision and natural language tasks. \citet{uhlich2020mixed} proposed a re-parametrization that allows to select the bitwidth for each layer through first order optimization, while also relying on STE. The re-parametrization is more complex than the additive noise used in \diffq, and suffers from the biased gradient of STE. Results suggest that \diffq achieves similar or better trade-offs between model size and accuracy. Besides, in the present work we explore setting a bitwidth for individual groups of weights within each layer, rather than layer-wise.

The limitations of STE methods for quantization were first noticed by \citet{liu2019learning}. They recommend using a linear combination of the unquantized and quantized weight, with the gradient flowing only through the unquantized contribution. In a similar spirit, \citet{fan2020training} sample for each layer and iteration whether to use the quantized or unquantized weight. Both methods reduce the bias from STE, but also remove some of the quantization noise during training. In contrast our method allows to keep a full pseudo quantization noise without the STE bias. \citet{liu2022nonuniform} proposed the \emph{Generalized STE} method to deal with gradient instabilities by calculating the expectation of the stochastic quantization during the backward phase. Finally, \citet{nagel2022overcoming} extend the analysis we present in Section~\ref{sec:counter} on the oscillations of weights when using STE and suggest tracking the weight oscillations in order to freeze them when needed, as an ad-hoc solution.

A last line of related work is Product Quantization (PQ)~\cite{stock2019and}, where code words are being learned to quantize blocks of weights rather than single weights. This method achieves a higher compression level than per-weight quantization but also requires carefully
choosing the size of the codebooks for each layer. In contrast, our method requires only choosing a single hyper-parameter to balance between model size and accuracy. Besides, as noted by~\citet{fan2020training}, per-weight quantization and PQ can be combined. We compare with PQ on vision and language tasks: while PQ can reach smaller model size than \diffq, it can also suffer from unacceptable accuracy loss, in particular for language modeling.
\section{Background}
\label{setup}

Let us consider a weight vector $w \in \real^d$, where $d\in\nat$, typically the weights of convolution or linear layer.
Each entry of the vector is typically coded over 32 bits with floating-point precision. We aim to reduce the number of possible states to $2^B$, where $B \ll 32$ is the number of bits of precision. First, we assume $w_i \in [0, 1]$ for all $1 \le i \le d$. In practice, one would first normalize $w$ as
\[
\hat{w} = \displaystyle\frac{w - \min(w)}{\max(w) - \min(w)},
\] 
and provide the tuple  $(\min(w), \max(w))$ separately as a 32 bits IEEE float. Given that for typical deep learning models $d \gg 1$, storing this range has a negligible cost. For readability, we describe the method for scalar values $w \in [0, 1]$, however, this can be easily extended to vectors $w\in\real^d$.

\subsection{Uniform quantization}

The simplest quantization methods consist of taking $2^B$ points evenly spaced in the range $[0, 1]$ and round each entry of $w$ to the nearest point. One can then store the rounded value by its index, which requires only $B$ bits.
Formally, we quantize a number $w \in [0, 1]$ over $B$ bits as
\begin{equation}
\label{eq:uniform}
   \forall w\in[0, 1], B \in \nat_*, \Q(w, B) =  \frac{\mathrm{round}\left(w \cdot (2^{B} - 1)\right)}{2^{B} - 1}.
\end{equation}

While the intuitive definition of quantization is for an integer number of bits, we can extend the previous definitions to a real-valued number of bits $B \in \real_{*+}$.
Note that variants of this scheme exist, for instance, symmetric uniform quantization, which enforces that $0$ is always
exactly represented~\citep{krishnamoorthi2018quantizing}.

\subsection{Optimization of the quantized weights}

The weight vector $w$ is typically obtained through the process of training a predictor function parameterized by $w$, denoted as $f_w$, to minimize a loss function $L$, 
\begin{equation}
    \min_{w\in{\real^d}} L(f_w),
\end{equation}
where $L(f_w)$ is the empirical risk over a given dataset. The process of quantizing a vector $w$ over $B$ bits introduces a quantization noise $\N(w, B) = \Q(w, B) - w$, which is unaware of the training objective $L$. Even if $w$ is close to the optimum, $\Q(w, B)$ might deteriorate arbitrarily the performance of the predictor.

Thus, given a fixed budget of bits $B$, one would ideally like to minimize the empirical risk when considering the quantization process,
\begin{equation}
    \min_{w\in{\real^d}} L(f_{\Q(w, B)}),
\end{equation}
where $f_{\Q(w, B)}$ is the predictor function using the quantized model parameters.

Unfortunately, the gradients of $\Q(w, B)$ are zero over its definition domain because of the rounding operation, and as a result, it cannot be optimized using first-order optimization methods such as SGD or Adam~\citep{adam}. One possible solution is to replace the Jacobian of $\Q(\cdot, B)$ with the identity matrix during the backward phase, as suggested in the STE method~\citep{bengio2013estimating}. The STE method was popularized for quantization as the Quantization Aware Training (QAT) technique by~\citet{krishnamoorthi2018quantizing}.

\subsection{The instability and bias in STE}
\label{sec:counter}

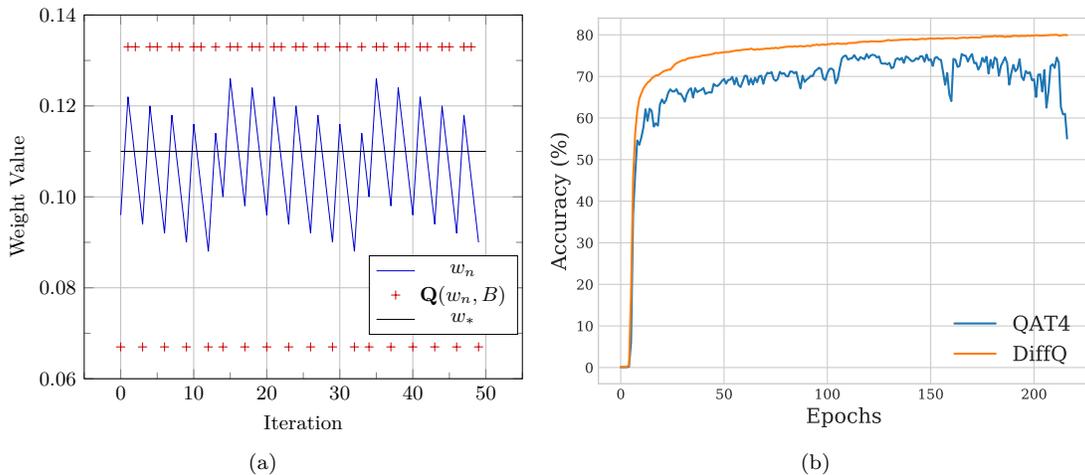
\begin{figure}[t!]
    \centering
    \subfigure[]{\label{}\begin{tikzpicture}[
	every node/.style={font=\fontsize{9}{5}\selectfont},
	scale=0.85,
]
\begin{axis}[
	xlabel={Iteration},
	ylabel={Weight Value},
	x label style={at={(axis description cs:0.5,-0.08)},anchor=north},
    y label style={at={(axis description cs:-0.11,.5)},rotate=0,anchor=south},
	legend entries={$w_n$, {$\Q(w_n, B)$}, $w_*$},
    legend style={
        overlay,
        font=\fontsize{9}{5}\selectfont,at={(0.99,0.23)},anchor=east,
        legend columns=1, fill=white,draw=black},
    ymin=0.06,
    ymax=0.14,
    ytick distance={0.02},
    ytick distance={0.02},
    minor x tick num=1,
    minor y tick num=1,
	grid=major,
	y tick label style={/pgf/number format/.cd,
		fixed,
		fixed zerofill,
		precision=2,},
]
\addplot [blue!80!black,
] coordinates {
(0, 0.096) (1, 0.122) (2, 0.108) (3, 0.094) (4, 0.120) (5, 0.106) (6, 0.092) (7, 0.118) (8, 0.104) (9, 0.090) (10, 0.116) (11, 0.102) (12, 0.088) (13, 0.114) (14, 0.100) (15, 0.126) (16, 0.112) (17, 0.098) (18, 0.124) (19, 0.110) (20, 0.096) (21, 0.122) (22, 0.108) (23, 0.094) (24, 0.120) (25, 0.106) (26, 0.092) (27, 0.118) (28, 0.104) (29, 0.090) (30, 0.116) (31, 0.102) (32, 0.088) (33, 0.114) (34, 0.100) (35, 0.126) (36, 0.112) (37, 0.098) (38, 0.124) (39, 0.110) (40, 0.096) (41, 0.122) (42, 0.108) (43, 0.094) (44, 0.120) (45, 0.106) (46, 0.092) (47, 0.118) (48, 0.104) (49, 0.090)
};

\addplot [red!80!black, mark=+, only marks
] coordinates {
(0, 0.067) (1, 0.133) (2, 0.133) (3, 0.067) (4, 0.133) (5, 0.133) (6, 0.067) (7, 0.133) (8, 0.133) (9, 0.067) (10, 0.133) (11, 0.133) (12, 0.067) (13, 0.133) (14, 0.067) (15, 0.133) (16, 0.133) (17, 0.067) (18, 0.133) (19, 0.133) (20, 0.067) (21, 0.133) (22, 0.133) (23, 0.067) (24, 0.133) (25, 0.133) (26, 0.067) (27, 0.133) (28, 0.133) (29, 0.067) (30, 0.133) (31, 0.133) (32, 0.067) (33, 0.133) (34, 0.067) (35, 0.133) (36, 0.133) (37, 0.067) (38, 0.133) (39, 0.133) (40, 0.067) (41, 0.133) (42, 0.133) (43, 0.067) (44, 0.133) (45, 0.133) (46, 0.067) (47, 0.133) (48, 0.133) (49, 0.067)
};

\addplot[black] coordinates { (0, 0.11) (50, 0.11)};
\end{axis}
\end{tikzpicture}}
    \subfigure[]{\label{fig:app_oscillations}\includegraphics[width=0.45\columnwidth]{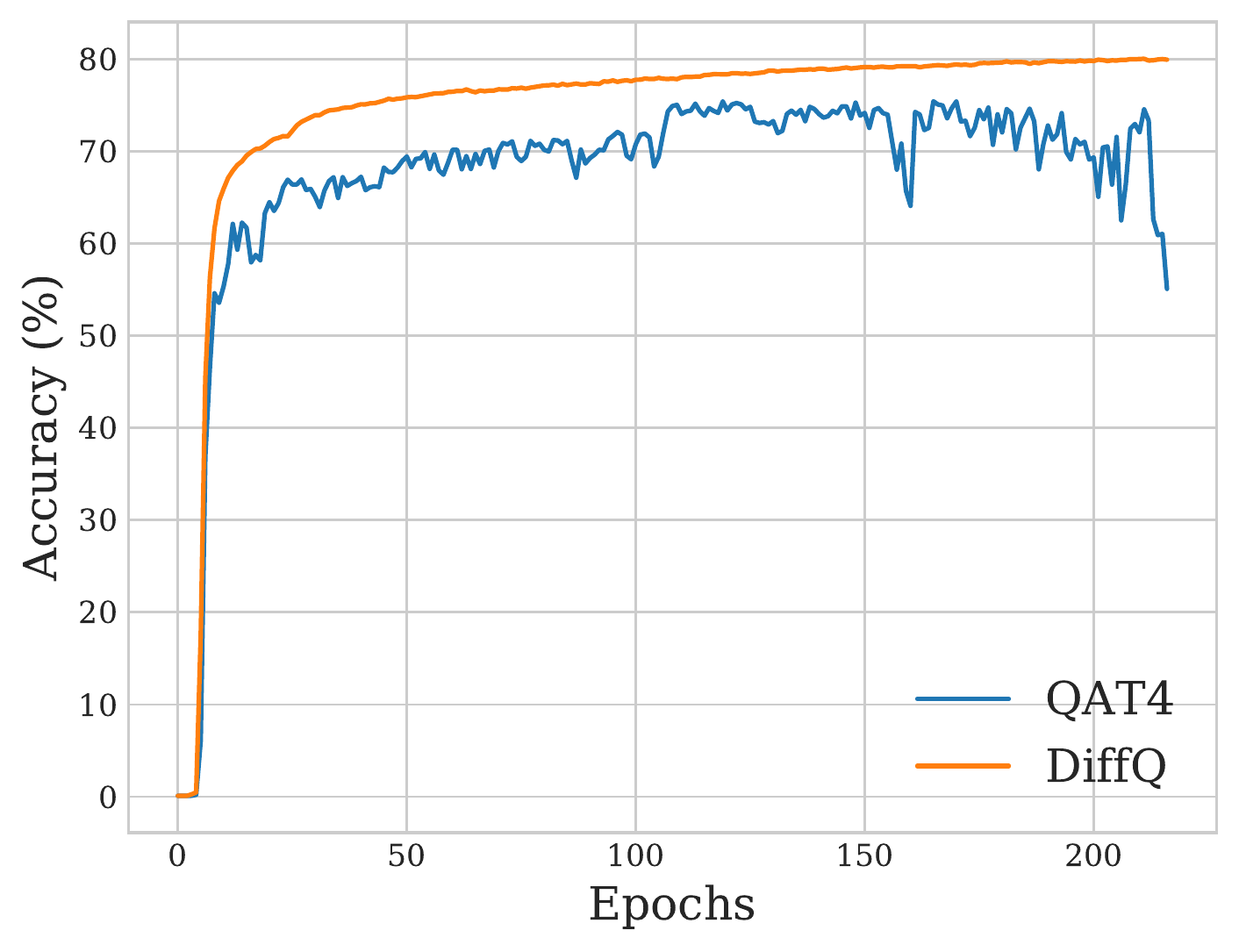}}
    \caption{
    \looseness=-1
        \textbf{(a)} Using STE and SGD to optimize the 1D least-mean-square problem given by \eqref{counter_example_ste} (with $B=4$ and $X=1$ a.s.). $\Q(w_n, B)$ oscillates between the quantized value just above ($w_+$) and just under ($w_-$) the unquantized
        ground truth $w_*$, while $w_n$ oscillates around the boundary $(w_+ + w_-) / 2$. \textbf{(b)} Model accuracy vs. epochs for ImageNet using EfficientNet-b3. Results are presented for both QAT over 4 bits and \diffq.
    }
    \label{fig_counter_example}
    \vskip 0.1in
\end{figure}

As described by~\citet{fan2020training}, following the STE approach can cause instability during training and bias in the models' gradients and weights. As a result optimization will fail to converge to the optimal value even on simple cases. To demonstrate that, consider the following 1D least-mean-square problem, where $B\in \nat_*$, the optimal weight $w_* \in [0, 1]$ such that $\Q(w_*, B) \neq w_*$, and $\Q(w_*, B) \in (0, 1)$. Given a random variable $X \in \real$ with $\sigma^2 = \esp{X^2}$ such that $0 < \sigma^2 < \infty$, we would like to minimize the following using STE based QAT:
\begin{equation}
\label{counter_example_ste}
    \min_{w\in[0, 1]} L(w) := \esp{\frac{1}{2}\left(X\Q(w, B) - X w_*\right)^2}.
\end{equation}
We immediately have that the optimum is achieved for $\Q(w, B) = \Q(w_*, B)$.
Let us try to optimize \eqref{counter_example_ste} using SGD with STE starting from $w_0 = w_*$, with $w_n$ the sequence of iterates.
We call $w_-$ and $w_+$ the quantized values just under and above $w_*$, and we assume without loss of generality that $\Q(w_*, B) = w_+$. The expected gradient with STE at iteration $n$ is given by
\begin{equation}
\label{counter_grad}
    G_n = \sigma^2 (\Q(w_n, B) - w_*).
\end{equation}
In particular, $G_0 = \sigma^2 (w_+ - w_*) > 0$, and $G_n$ will stay positive until $\Q(w_n, B) = w_-$. At this point, we will have $G_n < 0$,
and will stay so until $\Q(w_n, B) = w_+$. Thus, we observe that using STE, $\Q(w_n, B)$ will oscillate between $w_-$ and $w_+$, while the optimal value is $w_+$. The pattern of oscillation will depend on the learning rate and relative position of $w_*$ within the segment $[w_-, w_+]$. Taking a smaller step size will reduce the amplitude of the oscillations of $w_n$, but not of $\Q(w_n, B)$, which is what interests us. Indeed, $w_n$ oscillations are centered at the boundary $(w_+ + w_-) / 2$.
We provide one example of those oscillations on Figure~\ref{fig_counter_example} with $w_*=0.11$, $B=4$, $X=1$ a.s. and a step size of 0.5.

Extrapolating to a model with millions of parameters, at any point in time, a significant fraction of the weights could be quantized to a suboptimal value due to the oscillations implied by the STE method.
We conjecture that this behavior explains the oscillations of the accuracy observed when training an EfficientNet-b3 with QAT using 4 bits per weight on ImageNet (see Figure~\ref{fig:app_oscillations}).
In the following section, we introduce \diffq, a method based on independent additive pseudo quantization noise, that does not suffer from such a bias, while approximating well enough quantization noise to perform efficient quantization aware training.

\section{Method}
\label{method}

\paragraph{Pseudo quantization noise.}
A classical assumption in digital signal processing when working with quantized signals is that the quantization noise is approximated by independents uniform
variables over $[-\Delta/2, \Delta/2]$ with $\Delta = \frac{1}{2^B - 1}$ the quantization step. This approximation was studied in depth by~\cite{widrow1996statistical}
as Pseudo Quantization Noise (PQN). Following this assumption, we define the \emph{pseudo quantization function} $\tQ$
for all $x \in \real$ and $B \in \real_{+*}$ as
\begin{equation}
\label{eq:diffq}
    \tQ(x, B) = x + \frac{\Delta}{2} \cdot \mathcal{U}[-1, 1],
\end{equation}
with $\mathcal{U}[-1, 1]$ an independent sample from the uniform distribution over $[-1, 1]$.
This pseudo quantization function is differentiable with respect to $x$ and $B$.
Unlike QAT, this differentiability does not require an STE. It also provides a meaningful gradient with respect
to the number of bits used $B$ (extended to be real-valued).

If we look back at the example from Figure~\ref{fig_counter_example}, using now \eqref{eq:diffq} instead of STE, the expected gradients for SGD become
\begin{align}
\nonumber
G_n &= \esp{x \cdot \left(\left(w_n + \frac{\Delta}{2} \cdot \mathcal{U}[-1, 1]\right) x - w_* x\right)} \\
	&= \sigma^2 (w_n - w_*),
\end{align}
which cancels out for $w_n = w_*$, so that at convergence we indeed have $\Q(w_n, B) = \Q(w_*, B)$, i.e. the gradient estimate is unbiased
and converges to the right solution.

\paragraph{Mixed precision.}
We used a common precision $B$ for all the entries of the weight vector $w$. One can instead use different values for different entries.
Formally, the entries in $w$ are grouped by considering $w\in\real^{g\times d/g}$ with $g$ the group size and $d / g$ the number of groups.
We can then extend the definition of $\Q(w, B)$ given by \eqref{eq:uniform} and \eqref{eq:diffq} to use a number of bits $b_s$ for the group $s$, with $b \in \real{*+}^{d/g}$.

\paragraph{Training objective.}
Given $w \in \real^{g \times d /g}$ with $g$ groups of $d/g$ entries, and a number of bits $b \in \mathbb{N}_{*}^g$, we define the model size, expressed in MegaBytes ($1\mathrm{MB} = 8\cdot 2^{20}$ bits)
\begin{equation}
\label{eq:model_size}
\M(b) = \frac{g}{2^{23}}\sum_{s=1}^{d/g} b_s.
\end{equation}
A typical objective of quantization is to achieve the best possible performance within a given model size budget or to achieve the smallest model size
that reaches a given performance, i.e. we want to minimize with $b \in \nat_*^{d/g}$, and $w\in\real^{g\times d/g}$ either,
\begin{equation}
\begin{aligned}
     & \min_{w, b} L(f_{\Q(w, b)}),\\
    &\,\text{s.t.} \quad \M(b) \leq m.
\end{aligned}
\qquad \text{or} \qquad
\begin{aligned}
     & \min_{w, b} \M(b),\\
    &\,\text{s.t.} \quad L(f_{\Q(w, b)}) \leq l.
\end{aligned}
\end{equation}
We can relax $b$ to be real valued, and replace $\Q$ by our differentiable pseudo quantization function $\tQ$. Then, following the \emph{exact penalty method}~(\citet{bertsekas1997nonlinear}, Section~4.2, \citet{bertsekas2014constrained}, Chapter 4), 
there is $\lambda(m) > 0$ (or $\lambda(l)$ for the right hand side problem), such that the left hand size problem is equivalent to
 \begin{equation}
 \label{eq:opt_diffq}
 \min_{w, b} L(f_{\tQ(w, b)}) + \lambda(m) \M(b),
 \end{equation}
 which is fully differentiable with respect to $w$ and $b$ and can be optimized with first order optimization.

\paragraph{Parametrization.}
 In practice, the number of bits used for each group $b\in \real^g_{*+}$
 is obtained from a logit parameter $l \in \real^g$, so that we have
 \begin{equation}
     b = b_\textrm{min} + \sigma(l) (b_\textrm{max} - b_\textrm{min}),
 \end{equation}
 with $\sigma$ is the sigmoid function, and $b_\textrm{min}$ and $b_\textrm{max}$ the minimal and maximal number of bits to use. The trainable parameter $l$ is initialized so that $b = b_\textrm{init}$. We set $b_\textrm{init} = 8$.
 
\paragraph{Evaluation and noise distribution.}
 At evaluation time, we round the value $b$ obtained from \eqref{eq:opt_diffq} as $\tilde{b} = \mathrm{round}(b)$ and quantize $w$ as $\Q(w, \tilde{b})$.
 Thus, the amount of quantization noise at evaluation can be larger than the amount of noise injected at train time. We observed that using a noise distribution with larger support, such as Gaussian noise with unit variance (i.e. 3 times the variance of $\mathcal{U}([-1, 1])$), makes the model more robust to this operation. An empirical comparison between uniform and Gaussian noise can be found in Table~\ref{supp:noise} in the Appendix. Thus in the rest of the paper, we always use Gaussian noise at train time.

 \paragraph{True model size.}
 The mode size given by \eqref{eq:model_size} is used at train
 time but does not account for part of the true model size.
 At evaluation time, we represent each weight by the integer obtained from the rounding operation in \eqref{eq:uniform}. For each layer in the network, we also store two 32 bits float numbers for the minimum and maximum scale. Finally, the actual value of $\tilde{b}$ must be coded, as it is no longer a fixed constant. For each layer, we compute the maximum value of $C_s = \log_2(1 + \tilde{b}_s - b_\mathrm{min})$ over all groups $s \in \{1, \ldots, d / g\}$. We encode once the value $\max(C)$ as an 8-bit integer, and for each group, we encode $b_s - b_\mathrm{min}$ over $\max(C)$ bits. The true size for one layer, expressed in MegaBytes, is thus given by
 \begin{equation}
 \label{eq:true_model_size}
     \hspace{-0.1cm}\tilde{\M}(b) =  \frac{1}{2^{23}}\left(2 \cdot 32 + 8 + \frac{d}{g} \max(C) + g\sum_{s=1}^{d/g} b_s \right).
 \end{equation}
\section{Results}
\label{results}
\renewcommand\UrlFont{\color{purple}\rmfamily}
We present experimental results for language modeling, audio source separation, and image classification. We show that \diffq{} can often provide a model with comparable performance to the uncompressed one while producing a model with a smaller footprint than the baseline methods (STE based).
We provide a finer analysis of different aspects of \diffq hyper-parameters and their impact on quantized models in next Section. Finally, we discuss limitations of DiffQ in the Limitation Section. Both experimental code, and a generic framework usable with any architecture in just a few lines, is available
on our Github \href{https://github.com/facebookresearch/diffq}{github.com/facebookresearch/diffq}.
All hyper-parameters for optimization and model definition are detailed in the Appendix. In all tables, $\uparrow$ (resp. $\downarrow$) indicates that highest is best (resp. lowest is best). All results referred to as ``QAT'' are obtained using the formula
given by \eqref{eq:uniform} with a layer-wise min-max scaling of the weights. When using \diffq, we use the same per layer min-max scaling. When also doing activation quantization, we use per-channel min-max scaling of the activations. All \diffq experiments use Gaussian noise as explained in Section~\ref{method}.

\begin{table}[t!]
\begin{center}
\begin{small}
\caption{Comparison of \diffq against baselines presented in the Related Work section. Sizes marked with $^\dagger$ are reported after Huffman coding, following~\citet{polino2018model}. Accuracies marked with $^*$ are the best rather than last one to match previous practices.}
\label{tab:related}
\centering
\begin{sc}
\resizebox{0.8\columnwidth}{!}{
\begin{tabular}{l|l|c|r}
\toprule
Model       & Method     & Top-1 Acc. (\%) & M.S. (MB)   \\ \midrule
\multicolumn{4}{c}{CIFAR10}                    \\ \midrule
ResNet-18   & Uncompressed      & \textbf{95.3}  & 42.7  \\ \midrule
ResNet-18   & UNIQ~\cite{baskin2018uniq}      & 89.1       & \textbf{2.7}  \\ 
ResNet-18   & NICE~\cite{baskin2018nice}      & 92.7       & \textbf{2.7}  \\ 
ResNet-18   & \diffq~(Ours)     & \textbf{93.9}       & \textbf{2.7}  \\ 
\midrule
ResNet-20   & Uncompressed      & \textbf{92.7}*  & 1.48  \\ 
\midrule
ResNet-20   & DQ~\cite{uhlich2020mixed}         & 91.4*      & 0.07   \\ 
ResNet-20   & \diffq~(Ours)       & \textbf{91.6*}      & \textbf{0.06}   \\ \midrule
\multicolumn{4}{c}{CIFAR100}                 \\ \midrule 
Wide-ResNet & Uncompressed    	            & \textbf{76.2}  & 139.4\\
\midrule
Wide-ResNet & DiffQuant~\cite{polino2018model}  & 49.3       & 7.9  \\ 
Wide-ResNet & \diffq~(Ours)       & \textbf{75.6}       & \textbf{4.7}  \\ \midrule
\multicolumn{4}{c}{ImageNet}                 \\ \midrule
ResNet-18   & Uncompressed                  & 70.9*       & 44.6  \\ 
\midrule
ResNet-18   & Meta-Quant~\cite{chen2019metaquant} & 60.3       & \textbf{1.3}  \\ 
ResNet-18   & DQ~\cite{uhlich2020mixed}         & 70.1*       & 5.4  \\ 
ResNet-18   & LSQ 4 bits~\cite{esser2020learned}        & 70.7*       & 5.6  \\ 
ResNet-18   & \diffq~(Ours)       & \textbf{71.1}*       & \textbf{5.3}  \\ 
\midrule
ResNet-50   & Uncompressed                       & \textbf{77.1}*       & 97.5  \\ 
\midrule
ResNet-50   & LSQ 4 bits~\cite{esser2020learned}        & 76.2*       & 12.3 \\ 
ResNet-50   & LSQ 3 bits~\cite{esser2020learned}        & 75.6*       & 9.3  \\ 
ResNet-50   & \diffq~(Ours)       & \textbf{76.6}*      & 10.5 \\ 
ResNet-50   & \diffq~(Ours)       & 76.3*       & \textbf{8.8}    \\ \bottomrule
\end{tabular}}
\end{sc}
\end{small}
\end{center}
\end{table}

\subsection{Comparison to related work}
\label{sec:compare_related}
On Table~\ref{tab:related}, we compare \diffq to some of the related work presented in Section~\ref{relatedwork}. Compared with the NICE~\citep{baskin2018nice} and UNIQ~\citep{baskin2018uniq} methods, which also rely on additive noise, \diffq achieves significantly better accuracy for the same model size. We then compare to the differentiable quantization method by \citep{polino2018model}, which only optimizes the non uniform quantization points, not the pre-quantization weights. Following their practice, we report numbers after Huffman coding. We achieve a model almost half as small, with a gap of 25\% in accuracy, proving that optimizing pre-quantization weights is more important than tuning a non uniform quantization grid. Meta-Quant~\citep{chen2019metaquant} achieves smaller model size than \diffq, with 1 bit per weight, a regime where the PQN assumption breaks down, at the price of losing nearly 10\% of accuracy. Finally, compared with two quantization methods: DQ by~\citet{uhlich2020mixed} and LSQ by~\cite{esser2020learned}. When considering DQ, \diffq achieves slightly smaller model size and better accuracy on ImageNet using ResNet-18, and a 15\% smaller model with sightly better accuracy for a Resnet-20 trained on CIFAR-10. Comparing to LSQ~\footnote{We used our own LSQ implementation, with only weight quantization, since no official code is available. Comparison with the results reported in~\cite{esser2020learned} can be found on Table~\ref{tab:imagenet_supp}.}, \diffq achieves better accuracy with smaller model size on ImageNet using both ResNet-18 and ResNet-50. Additional comparison between \diffq and LSQ for higher compression rates can be on Table~\ref{tab:imagenet_supp} in the Appendix.

\begin{table}[t!]
\caption{
Language modeling results for a 16 layer Transformer trained on Wikitext-103. We also test combining weight and activation quantization. We compared \diffq to QAT and Quant-Noise (QN) method proposed by~\citet{fan2020training} (models with $\dagger$ were trained with a layer-drop of 0.2~\cite{fan2019reducing}). Activations are quantized over 8 bits, with a per-channel scaling.}
\label{tab:text}
\begin{center}
\begin{small}
\begin{sc}
\resizebox{0.7\columnwidth}{!}{
\begin{tabular}{llcr}
\toprule
	Weights & Activation	 & PPL~$\downarrow$ & M. S. (MB)~$\downarrow$ \\
\midrule
Uncompressed & -  	   & \textbf{18.1} & 942\\
\midrule
    8 bits & 8 bits    & 18.3  & 236\\
QAT 8bits & 8 bits   	& 19.7  & 236 \\
QAT 4bits & 8 bits    	&   29.9   & 118 \\
LSQ 4 bits~\citep{esser2020learned} & 8 bits & 18.9 & 118 \\
\diffq  ($\lambda{=}5, g{=}16$) & 8 bits & \textbf{18.1}          & 130 \\
\diffq  ($\lambda{=}10, g{=}16$) & 8 bits & 18.6 & \textbf{113}\\
\midrule
Uncompressed $\dagger$ & -  	      & 18.3 & 942\\
\midrule
QN 8 bits$\dagger$~\cite{fan2020training}    & QN 8 bits & 18.7 & 236 \\
QN 4 bits$\dagger$~\cite{fan2020training}    & QN 8 bits & 19.5 & 118 \\ 
PQ$\dagger$~\cite{fan2020training}    & - & 20.7 & \textbf{38} \\ 
\bottomrule
\end{tabular}}
\end{sc}
\end{small}
\end{center}
\end{table}

\subsection{Language Modeling}
We trained a 16 layers transformer~\citep{vaswani2017attention} based language model on the Wikitext-103 text corpus~\citep{merity2016pointer},
following~\citet{baevski2018adaptive},
 and using the Fairseq framework~\citep{ott2019fairseq}. Results are presented in Table~\ref{tab:text}. We compare to the Quant-Noise method by~\citet{fan2020training}, but use a reduced layer-drop~\citep{fan2019reducing} of 0.1 instead of 0.2. 
 This both improves the baseline, as well as the performance of \diffq models. For \diffq, we explicitly set the gradient for the number of bits parameters to zero for all layers that have been dropped.
 In order to test the compatibility of \diffq with efficient int8 kernels, we further quantize the activations to 8 bits using PyTorch native support~\citep{Pytorch}.
 
 The transformer model has some tied parameters (e.g. word embedding in the first and pre-softmax layer). It is important to detect such tied parameters with \diffq. We use a single shared bits parameter when a parameter tensor is reused multiple times, and for each forward, we sample a single pseudo quantization noise per group of shared weights and reuse it appropriately. Failure to do so led to a significant worsening of the performance at validation time.
 
 While QAT breaks down when trying to get
 to 4 bits precision (perplexity of 29.9), using \diffq allows to achieve a lower model size (113MB vs. 118 MB 
 for QAT 4 bits) with a perplexity closer to the uncompressed one (18.6, vs.
 18.1 uncompressed). 
 We also tried fine-tuning a pre-trained model with LSQ \citep{esser2020learned}.
 While this works better than QAT, LSQ reaches a worst perplexity for a slightly larger model size than \diffq
 (18.9 perplexity for 118 MB). Similarly, Quant-Noise~\citep{fan2020training} improves on QAT but performs worse than \diffq, even when using more than twice as many bits. With just 4.4 bits per weight on average, \diffq achieve the same perplexity as the baseline. We also compare to PQ~\citep{stock2019and}, as reported by~\citet{fan2020training}.
 While PQ achieves higher compression levels, with just 38MB, its perplexity is the worst of all methods.

\subsection{Music Source Separation}

\begin{table}[t!]
\caption{Music source separation results for the Demucs model~\citep{defossez2019music}.  We report Signal-to-Distortion Ration (SDR) together with Model Size (M.S.).}
\label{tab:speech}
\begin{center}
\begin{small}
\resizebox{0.5\columnwidth}{!}{%
\begin{tabular}{lcr}
\toprule
		 & \textsc{SDR} (dB) ~$\uparrow$ & \textsc{M. S. (MB)}~$\downarrow$\\
\midrule
\textsc{Uncompressed}  &  \textsc{6.31}   & \textsc{1014} \\
\midrule
\textsc{QAT 4bits}    &	\textsc{5.99}  & \textsc{130} \\
\textsc{QAT 5bits}   & 	\textsc{\textbf{6.27}}  & \textsc{162} \\
\diffq  ($\lambda{=}3\mathrm{e}{-}4$) & \textsc{\textbf{6.28}} & \textsc{\textbf{120}} \\
\bottomrule
\end{tabular}}
\end{small}
\end{center}
\end{table}

\looseness=-1
We use the Demucs architecture by~\citet{defossez2019music} with $64$ initial hidden channels. The model is trained on the standard MusDB benchmark~\citep{musdb},
for 180 epochs, and evaluated with the Signal-To-Distortion Ratio (SDR) metric~\citep{measures}. The unquantized model is 1GB. We compare \diffq{} with QAT training with either 5 or 4 bits, with the results presented in Table~\ref{tab:speech}. With 5 bits, QAT is able to replicate almost the same performance as the uncompressed model.
When trying to further compress the model to 4 bits per weight, QAT leads to a sharp decrease of the SDR, losing 0.3dB, for a 130MB model. \diffq{} achieves a model size of 120MB, with only
a drop of 0.03dB of SDR compared to the uncompressed baseline.

\subsection{Image Classification}
\label{sec:image_classif}
Next, we evaluated three image classification benchmarks: ImageNet~\cite{imagenet_cvpr09}, CIFAR-10 and CIFAR-100~\cite{krizhevsky2009learning}. For CIFAR-10 and CIFAR-100 results are reported for MobileNet-v1~\cite{howard2017mobilenets}, ResNet-18~\cite{he2016deep}, and Wide-ResNet with 28x10, depth and width levels respectively ~\cite{zagoruyko2016wide}. ImageNet results are reported using EfficientNet-B3~\cite{tan2019efficientnet} and DeiT-B~\cite{touvron2020training} models. More details regarding hyper-parameters and augmentations used can be found in the Appendix.

\begin{figure}[t!]
\centering
\subfigure[CIFAR10]{\label{fig:cifar10}\includegraphics[width=0.47\columnwidth]{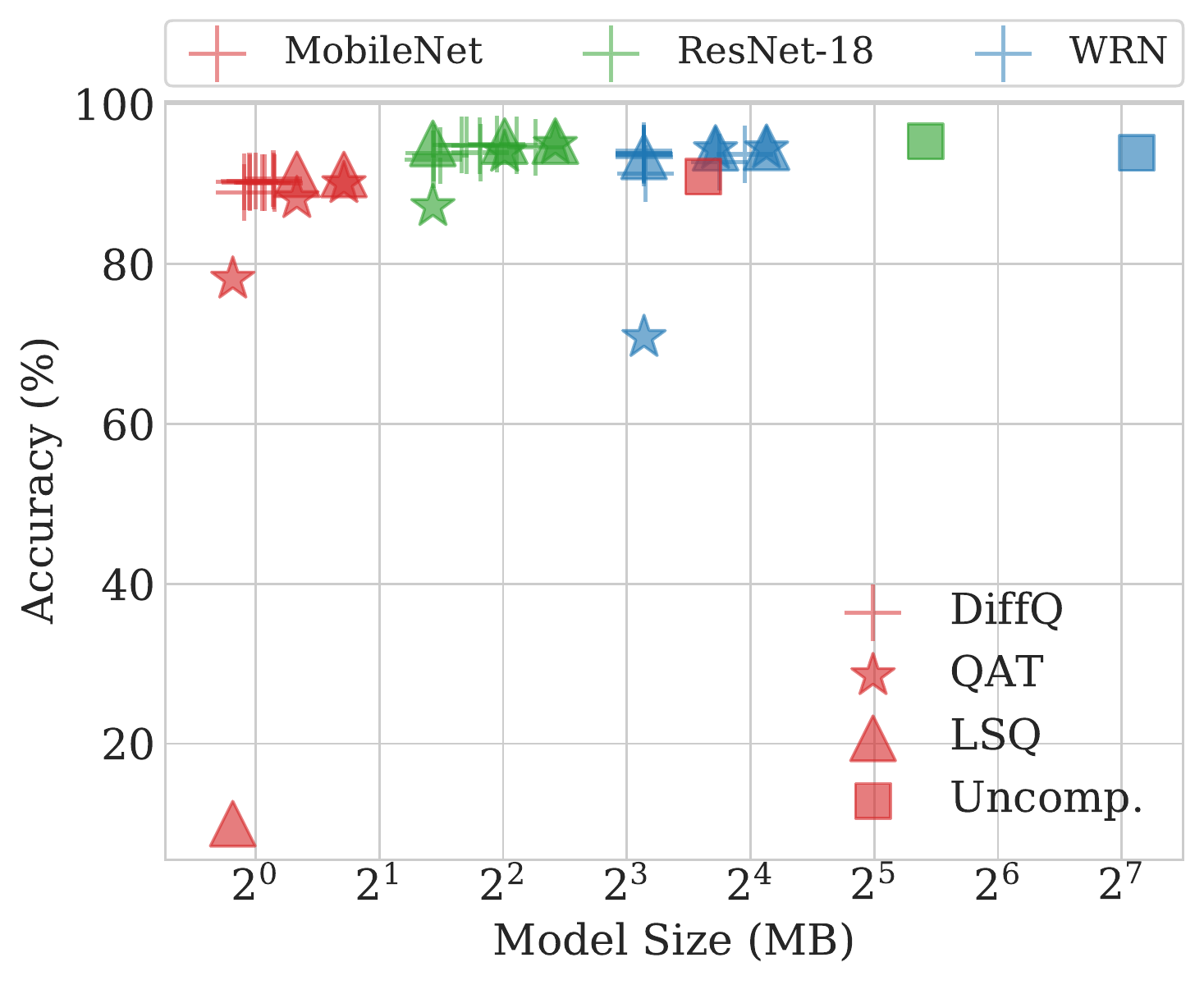}}
\subfigure[CIFAR100]{\label{fig:cifar}\includegraphics[width=0.46\columnwidth]{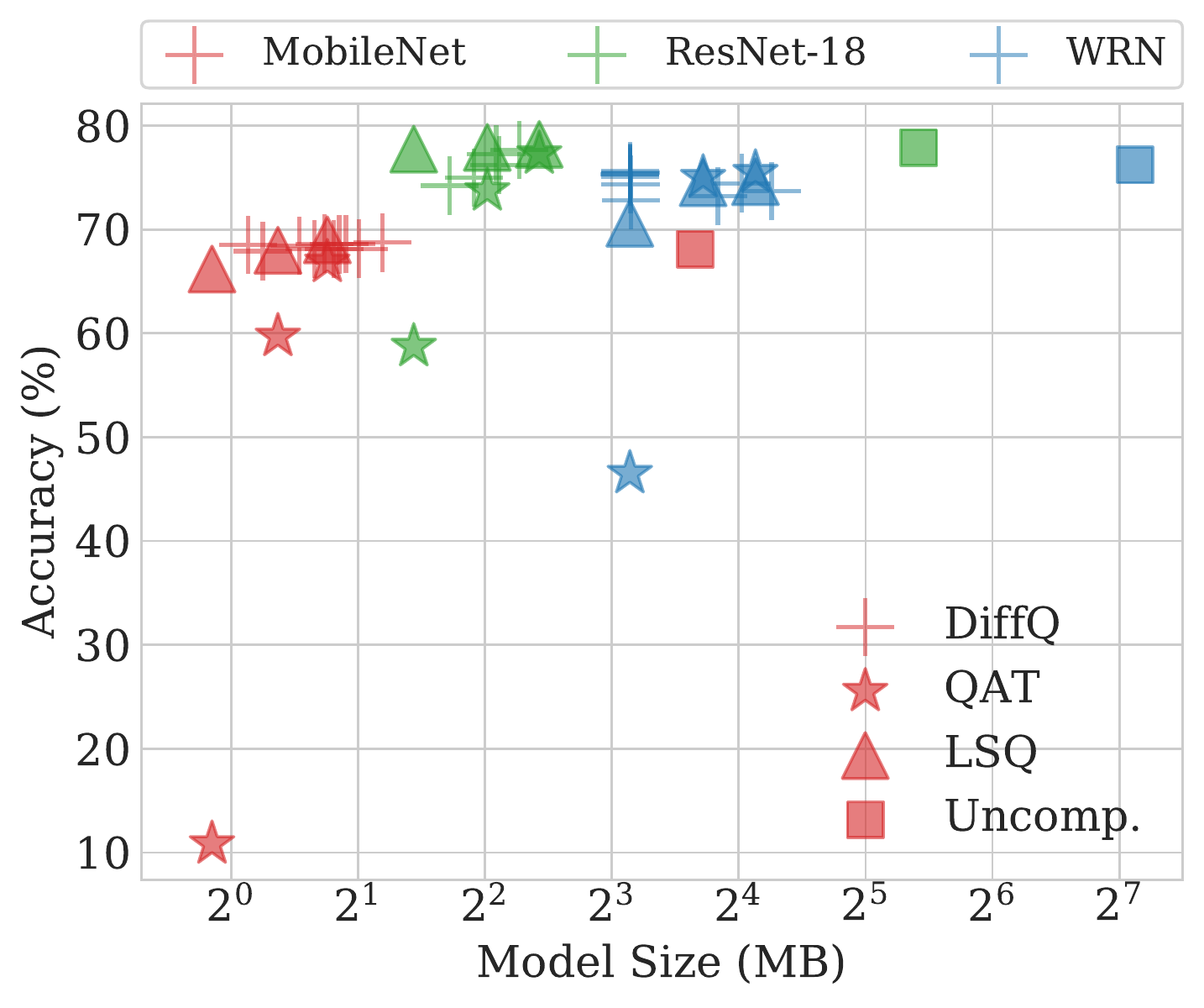}}
\caption{Model accuracy and size on CIFAR10 {\bf (a)} and CIFAR100 {\bf (b)} using MobileNet, ResNet-18, and WideResNet (WRN) models for various penalty levels using \diffq, QAT, LSQ, and the baseline.}
\label{fig:vision}
\end{figure}

\paragraph{CIFAR10 \& CIFAR-100.}
Results for CIFAR10 and CIFAR100 are depicted in Figures~\ref{fig:cifar10} and~\ref{fig:cifar}. We compare \diffq, QAT and LSQ (without activation quantization) using 2, 3, and 4 bits quantization. Performance of the uncompressed model is additionally presented as an upper-bound. 
To better understand the effect of the penalty level $\lambda$ on both model size and accuracy, we train models with \diffq using different penalty levels. 
Exact results are presented in Table~\ref{tab:cifar_supp}, in the Appendix, together with a detailed analysis.

Results suggest \diffq models reach comparable performance to the LSQ and outperforms QAT models while producing models with a smaller footprint. When considering 2 bits quantization, QAT is always worse than both LSQ and \diffq. While LSQ works well for Resnet18, it suffers from large drops in accuracies for MobileNet and WideResNet, failing entirely to train for MobileNet on CIFAR10, despite initialization from a pre-trained model.

\paragraph{ImageNet - DeiT.} Results for ImageNet using DeiT-B model are presented in Table~\ref{tab:imagenet}. We compared \diffq to QAT when training with 4 and 8 bits. Both QAT with 8 bits and \diffq reach comparable performance to the uncompressed model, while \diffq yields a model almost half of the size as QAT, however still bigger than QAT with 4 bits. When we increase $\lambda$, we get a smaller model-size than QAT with 4 bits but with better accuracy levels.

\begin{table}[t!]
\caption{Image classification results for the ImageNet benchmark. Results are presented for \diffq and QAT using 4 and 8 bits using the DeiT model~\citep{touvron2020training}. We report Top-1 Accuracy (Acc.) together with Model Size (M.S.).}
\label{tab:imagenet}
\begin{center}
\begin{small}
\begin{sc}
\resizebox{0.5\columnwidth}{!}{%
\begin{tabular}{l|cc}
\toprule
		& Top-1 Acc. (\%)~$\uparrow$ & M.S. (MB)~$\downarrow$ \\
\midrule
Uncompressed    	                  & 81.8 		  & 371.4 \\
\midrule
QAT 4bits    		                  & 79.2          & 41.7  \\
QAT 8bits    		                  & 81.6          & 82.9 \\
\midrule
\diffq ($\lambda{=}1\mathrm{e}{-}2$)  & \textbf{82.0} & 45.7  \\
\diffq ($\lambda{=}0.1$)  	          & 81.5          & \textbf{33.02}  \\
\bottomrule
\end{tabular}}
\end{sc}
\end{small}
\end{center}
\end{table}

\paragraph{ImageNet - EfficientNet.} We evaluate the performance of \diffq{} on the memory-efficient EfficientNet-B3 model. Results are depicted on Figure~\ref{fig:supp} (c) as well as in Table~\ref{supp:imagenet}, both in the Appendix. Both QAT 8 bits and \diffq{} achieves similar accuracy (QAT 81.3 \%, \diffq 81.5\%)
but with a smaller model size for \diffq (8.5MB vs. 12MB for QAT). 
When considering QAT 4 bits, \diffq{} produces a smaller
model with a significantly better accuracy level  (80.8\%). For QAT 4, we
noticed considerable instability close to the end of the training, see Figure~\ref{fig:supp} (b) in the Appendix.
\subsection{Analysis}
\label{analysis}

\begin{figure}[t!]
\centering 
\subfigure[EfficientNet-B3]{\label{fig:a}\includegraphics[width=0.49\columnwidth]{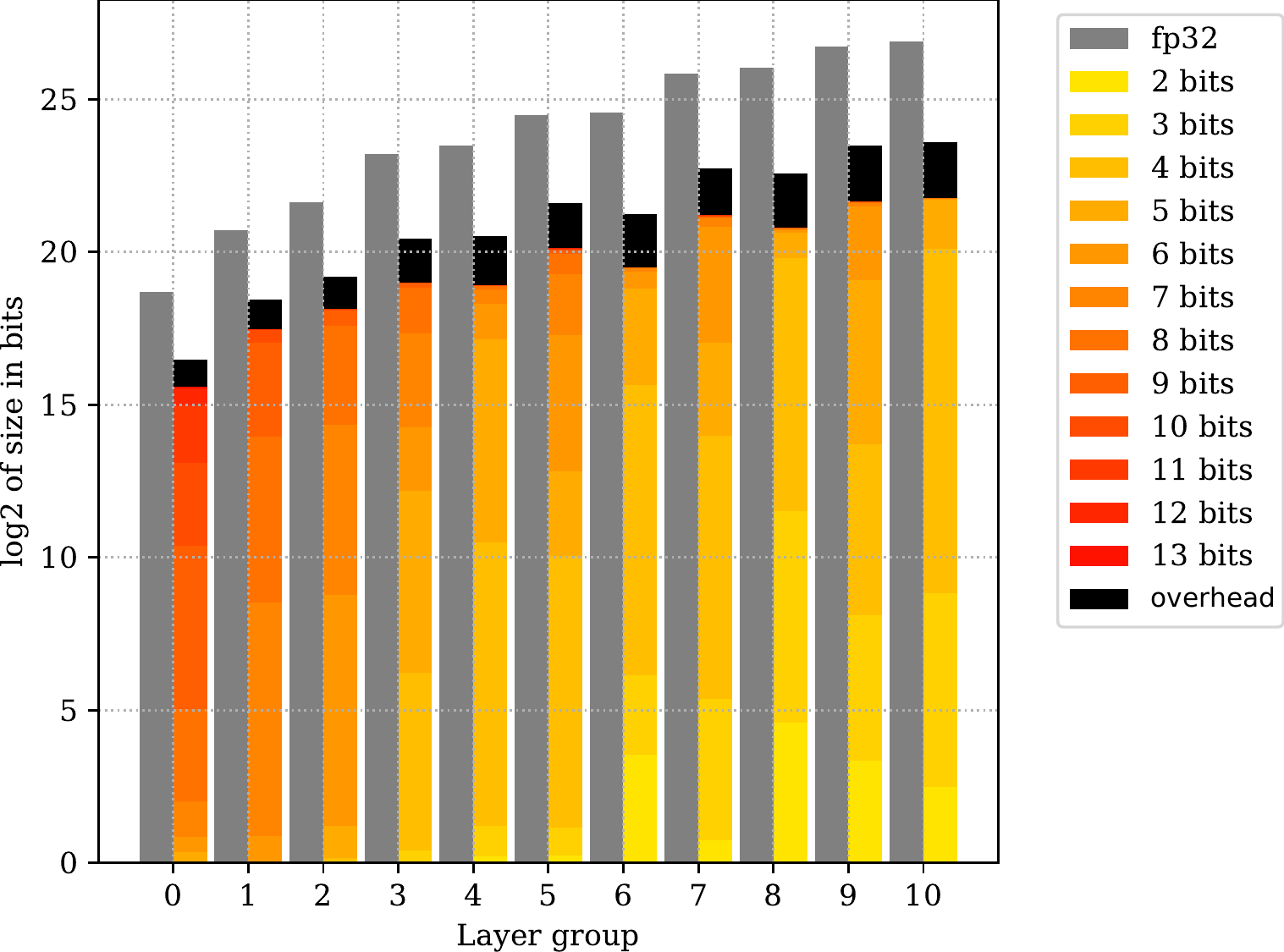}}
\subfigure[DeiT]{\label{fig:b}\includegraphics[width=0.49\columnwidth]{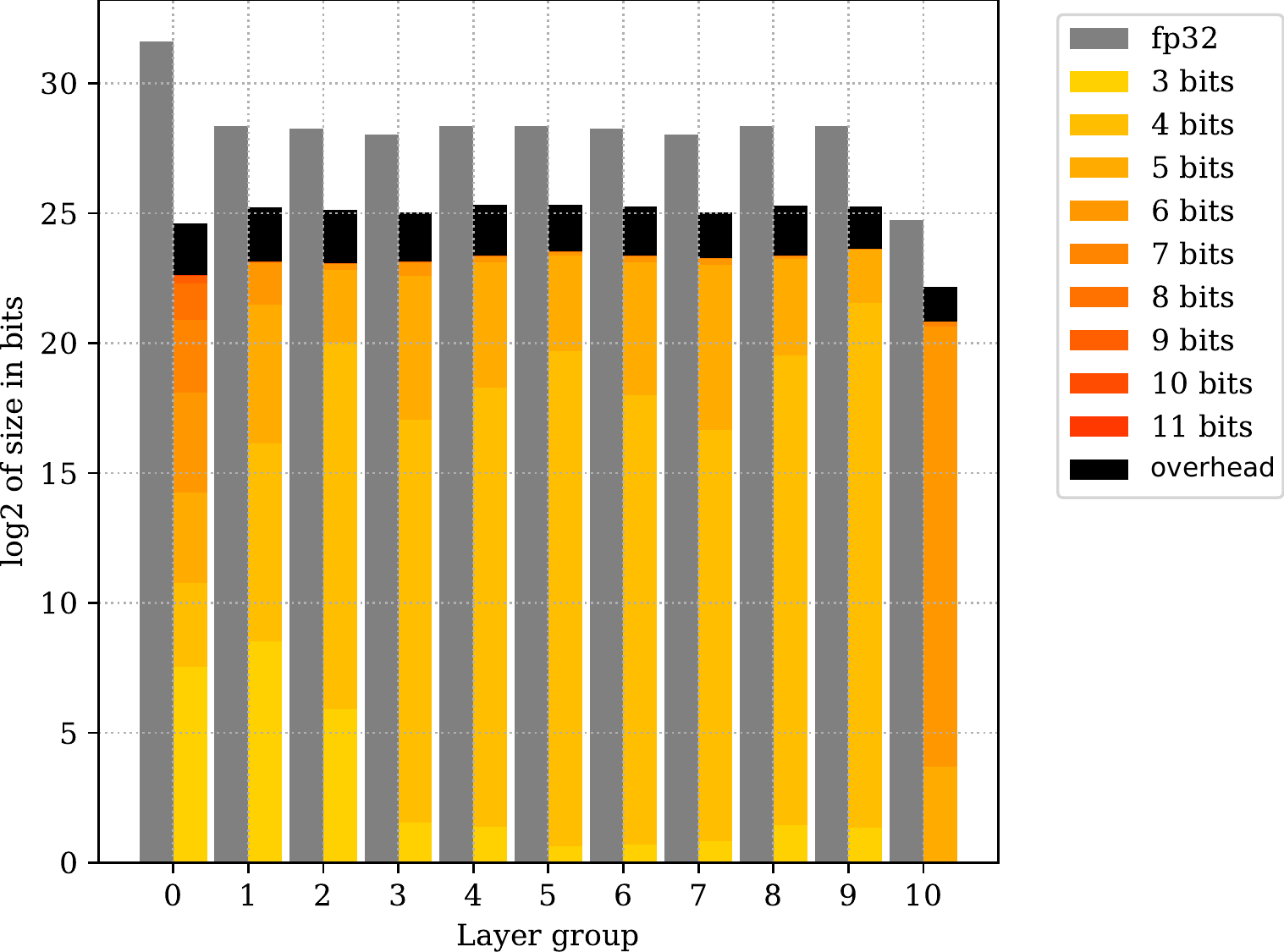}}
\caption{We group layers of a given architecture into 11 groups (group 0 being closest to the input, and 10 closest to the output), and report for
each group its contribution to the model size.
We compare the baseline EfficientNet-B3 (above) and DeiT (below) models (floating point 32 bits) and the quantized models with \diffq ($\lambda{=}5\mathrm{e}{-}3$ for EfficientNet-B3, $\lambda{=}1\mathrm{e}{-}2$ for DeiT). For quantized model, we also
report the distribution over each bitwidth within each group of layers. Scale is logarithmic across layers, and linear inside each one.
Finally, ``overhead'' shows the capacity needed to encode the bitwidth used for each group of weights.
}
\label{fig:hist}
\end{figure}

\paragraph{Bits Histogram.}
Figure~\ref{fig:hist} presents the weight bitwidth assignment over layer groups for the EfficientNet-B3 \cite{tan2019efficientnet} and DeiT \cite{touvron2020training} models trained on ImageNet. The capacity distribution over depth for ConvNets (EfficientNet-B3) and Transformers (DeiT) are different (fp32 shows uncompressed capacity). Notice, that the quantization trends are different too: for the ConvNet, smaller bitwidths are used for deeper layers of the model while large bitwidth is more common in the first layers (except for the last linear layer which seems to need some precision). For the Transformer, this effect of varying quantization by layer is similar but less pronounced, due to the more symmetric nature of the architecture.

\textbf{Fixed bitwidth.}
On  Table~\ref{supp:tab:app_image_abb} in the Appendix, we compare QAT to \diffq using a fixed number of bits, i.e. comparing strictly PQN to STE. On MobileNet, ResNet-18, and WideResNet for both CIFAR10 and CIFAR100, \diffq outperforms QAT, with a gap especially noticeable for 2 bits models, a regime where QAT becomes unstable, as we described in previous section.

\paragraph{Group size.}
\looseness=-1
We additionally evaluate the affect of the group-size, $g$, on model size and accuracy, by optimizing \diffq models using $g \in \{1, 4, 8, \infty \}$. When $g{=}\infty$, we use a single group for the entire layer. Results for ResNet-18 using CIFAR-100 are depicted in Figure~\ref{fig:gs} in the Appendix. Interestingly, we observed that increasing $g$, yields in a smaller model size on the expense of a minor decrease in performance. However, when setting $g{=}\infty$ model performance (model size and accuracy) is comparable to $g{=}8$ for this task.

\paragraph{Runtime overhead and loading time.}
Using DiffQ usually increase the training time by some amount.
On the language modeling task, the time per batch went from 115ms to 125ms.
When training a ResNet18 on CIFAR-10, it increased from 120ms to 150ms. For the Demucs model, it went from 0.9s to 1.1s. However, when training the EfficientNet-b3 model, we observed that the time per batch would nearly double. Thus it seems that for most architectures
the training time overhead is limited, although the worst case can be up to twice as slow.
At evaluation time, decompressing the Demucs model from its variable bitwidth compact representation takes around 2.81 seconds 
on a MacBook Pro with 2.4 GHz 8 cores Intel i9 processor.

\subsection{Limitations}
\label{limitations}

The model size given by~\eqref{eq:true_model_size} is obtained with a traditional encoding of the quantized model.
However, more efficient coding techniques exist when the entropy of the data is low, such as Huffman coding~\citep{huffman}. Using the ZLib library, we obtain an estimate of the Huffman compressed model size after quantization. For instance, for the language model described in Table~\ref{tab:text}, the QAT 8 model gets further compressed from 236MB to 150MB, showing that the entropy of its quantized weight is significantly lower than the maximal one for 8 bits integers. However, the \diffq model naive size is 113MB, and after compression by ZLib, gets to 122MB. This is a sign that the entropy is close to its maximal value, with ZLib adding only overhead for no gain.
In \eqref{eq:opt_diffq}, we only penalize the naive number of bits used, while asking for
the best possible accuracy. In that case, the model maximally use the entropy capabilities for a given number of bits. An interesting line of research would be to replace the model size \eqref{eq:model_size} to account for the actual entropy of the data, for instance with differentiable kernel density estimation. We leave that for further research.

Another limitation of DiffQ is that it can make training up to twice as slow, due to the extra parameters to optimize for and the more complex gradient calculation graph. Besides, in order to achieve a specific model size or accuracy, one has to tune the $\lambda$ penalty parameter. 

\section{Discussion}
\label{discussion}

\looseness=-1
We presented \diffq, a novel and simple differentiable method for model quantization via pseudo quantization noise addition to models` parameters. Given a single hyper-parameter that quantifies the desired trade-off between model size and accuracy, \diffq can optimize the number of bits used for each trainable parameter or group of parameters during model training. 
We conduct expensive experimental evaluations on various domains using different model architectures. Results suggest that \diffq is superior to the baseline methods on several benchmarks from various domains. On ImageNet, Wikitext-103, and MusDB, we achieve a model size that is smaller than a 4 bits quantized model, while retaining the same performance as the unquantized baseline.
For future work, we consider adapting the model size penalty to account for Huffman encoding, which could allow to further reduce the model size when it is gzipped. Another line of work would be using PQN to improve activation quantization, enabling 4-bits kernels for a larger number of tasks.

\bibliography{refs}
\bibliographystyle{tmlr}

\appendix
\renewcommand{\thesection}{\Alph{section}}
\onecolumn
\thispagestyle{empty}
\noindent\makebox[\linewidth]{\rule{\textwidth}{1pt}}
\begin{center}\Large \setstretch{1.2} Supplementary Material for \\
\textbf{Differentiable Model Compression via Pseudo Quantization Noise}
\noindent\makebox[\linewidth]{\rule{\textwidth}{1pt}}
\end{center}
\setcounter{section}{0}
\numberwithin{equation}{section}
\counterwithin{table}{section}
\counterwithin{figure}{section}
\setcounter{secnumdepth}{1}
\renewcommand\theHsection{\Alph{section}}

We provide in Section~\ref{app:xps} all the details on the 
exact hyper-parameters, models, and datasets used for the results in Section~\ref{results} of the main paper.
Then, we provide supplementary results in Section~\ref{app:results},
in particular tables for the scatter plots given on Figures \ref{fig:teaser} and \ref{fig:cifar}.

\section{Detailed experimental setup}
\label{app:xps}

All experiments are conducted using NVIDIA V100 GPUs with either 16GB or 32GB RAM, depending on the applications
(with language modeling requiring larger GPUs) on an internal cluster.
For all models trained with QAT or \diffq, we do not quantize tensors with a size under 0.01 MB (0.1 MB for the DeiT model).

\subsection{\diffq{} hyper-parameters}

For all experiments, we use $b_\textrm{min}=2$, $b_\textrm{max}=15$,
$b_\textrm{init}=8$ and Gaussian noise. 
We observed on most models that taking $b_\textrm{min} < 2$ is
unstable, with the notable exception of Resnet-20.
We use a separate Adam optimizer~\citep{adam} for the logit parameters controlling the number of bits used, with a default momentum $\beta_1=0.9$ and decay $\beta_2=0.999$.
We use the default learning rate $\alpha=1\mathrm{e}{-}3$ for all task,
except language modeling where we use $\alpha=1\mathrm{e}{-}2$.
The remaining hyper-parameters are $\lambda$, the amount of penalty
applied to the model size, and $g$, the group size. When $g$ is not mentioned, it is set to the default value $g=8$, which we found to be the best trade-off between the model freedom and the overhead from storing the number of bits used for each group. 

\subsection{Music Source Separation}

We train a Demucs source separation model~\citep{defossez2019music} (MIT license) with a depth of 6 and 64 initial hidden channels, on the MusDB dataset~\citep{musdb}\footnote{\url{https://sigsep.github.io/datasets/musdb.html}}, which is released under mixed licensing\footnote{\url{https://github.com/sigsep/website/blob/master/content/datasets/assets/tracklist.csv}}. 
All the training details are exactly as in~\citep{defossez2019music}.

\subsection{Language Modeling}

We trained a 16 layers transformer~\citep{vaswani2017attention} based language model on the Wikitext-103 text corpus~\citep{merity2016pointer}\footnote{\url{https://blog.einstein.ai/the-wikitext-long-term-dependency-language-modeling-dataset/}} released under the CC-BY-SA license,
following~\citet{baevski2018adaptive},
 and using the \texttt{Fairseq} framework~\citep{ott2019fairseq}, released under the MIT license. 
 We used the hyper-parameters and the script provided by \citep{fan2020training} in the \texttt{Fairseq} repository\footnote{\url{https://github.com/pytorch/fairseq/tree/master/examples/quant_noise}}, however, and unlike what they mention in their paper, this script does not include layer drop~\citep{fan2019reducing}.
 For \diffq, we tried the penalty levels $\lambda$ in $\{1, 5, 10\}$, with group size $8$, as well as $\lambda = 10$ and $g = 16$.
 For LSQ, we used the same training hyper-parameters as \diffq, except we initialized the model
 to a pre-trained model and used a learning rate 10 time smaller for fine tuning. Without this initialization, LSQ was failing to get under 40 of perplexity.

\noindent{\bf Tied weights and \diffq.}
The model we trained was configured so that the word embedding in the first layer and the weight of the adaptive softmax are bound to the same value. It is important to detect such bounded parameters with \diffq, as otherwise, a different number of bits could be used for what is in fact, the very same tensor. Not only do we use a single bits logit parameter
when a parameter tensor is reused multiple times, but for each forward, we make sure that the pseudo quantization noise
is sampled only once and reused appropriately. Failure to do so led to a significant worsening of the performance
at validation time.

\subsection{Image classification}
\label{supp:image}

\noindent{\bf CIFAR10/100.}
On the CIFAR10/100 datasets, we train 3 different models: MobileNet-v1~\citep{howard2017mobilenets}, ResNet-18~\citep{he2016deep}, and a Wide-ResNet with 28x10 depth and width levels respectively~\citep{zagoruyko2016wide}.
All experiments are conducted on a single GPU with a batch size of 128, SGD with a learning rate of 0.1, momentum of 0.9,
weight decay of $5\mathrm{e}{-}4$. The learning rate is decayed by a factor of 0.2 every 60 iterations.
To generate Figure~\ref{fig:cifar}, we evaluated \diffq for $\lambda$ in $\{0.01, 0.05, 0.1, 0.5, 1, 5\}$
and the group size $g$ in $\{4, 8, 16\}$.

For LSQ~\citep{esser2020learned}, we initialize from a pre-trained model. We tested both training for 90 epochs with a cosine schedule as originally done, or keeping the original learning rate schedule, only reducing the initial learning rate from 0.1 to 0.01. The second option achieved better overall results and that is the one we report.
We also tried training with LSQ from a randomly initialized model, but that performed the worst of all approaches.

The dataset has been obtained from the \texttt{torchvision} package\footnote{\url{https://github.com/pytorch/vision}}.
The input images are augmented with a random crop of size 32 with padding of 4, and a random horizontal flip.
The RGB pixel values are normalized to mean 0 and standard deviation 1. We use the default split between
train and valid as obtained from the \texttt{torchvision} package.

\noindent{\bf CIFAR-10 - Resnet 20.}
We use the implementation of Resnet 20 from~\citet{Idelbayev18a}.
We train for 600 epochs with a batch size of 128, with a learning rate of 0.1, momentum of 0.9, weight decay of $2\mathrm{e}{-}4$, and decrease the learning rate by a factor of 10, every 200 epochs.
We quantize all parameters except biases, we set the minimum number of bits to $b_{\mathrm{min}} = 1$ as we observe this was stable for this particular task, and lower the maximum number of bits to $b_{\mathrm{max}} = 10$. We use a group size $g=16$ and a penalty $\lambda = 10$ and a learning rate of $2\mathrm{e}{-}4$, for the separate Adam optimizer used for the bits parameters, which allows stay stable while going under 2 bits per weight.

\noindent{\bf ImageNet.}
We train an EfficientNet as implemented by~\citep{rw2019timm} (Apache license),
as well as a DeiT vision transformer~\citep{touvron2020training} (MIT license) on the ImageNet dataset~\cite{imagenet_cvpr09}\footnote{\url{http://www.image-net.org/}}. We use the original dataset
split between train and valid. 
The images go through a random resize crop to 300px, a random horizontal flip, and pixel RGB values are normalized
to have zero mean and unit variance.

\noindent{\bf ImageNet - EfficientNet.}
We trained for 100 epochs, using RMSProp~\cite{tieleman2012lecture} as
implemented in the \texttt{timm} package\footnote{\url{https://github.com/rwightman/pytorch-image-models}}
with a learning rate of 0.0016, a weight decay of $1\mathrm{e}-5$ and a momentum of 0.9. The learning rate is decayed
by a factor of 0.9875 with every epoch. As a warmup, the learning rate is linearly scaled from 0 to 0.0016 over the first 3 epochs. Following~\citep{rw2019timm}, we evaluate with an exponential moving average of the weights of the model, with
a decay of 0.99985. We use the random erase augmentation from \citep{rw2019timm}, as well as cutmix~\citep{yun2019cutmix},
with a probability of 0.2 and parameter to the beta distribution of 0.2.
All the models are trained on 8 GPUs. For \diffq, we used the penalties $\lambda$ in 
$\{5\mathrm{e}{-}4, 1\mathrm{e}{-}3, 5\mathrm{e}{-}3, 1\mathrm{e}{-}2, 5\mathrm{e}{-}2, 0.1, 0.5\}$
and the default group size $g=8$.

\noindent{\bf ImageNet - DeiT.}
We use the official DeiT implementation by \citet{touvron2020training}\footnote{\url{https://github.com/facebookresearch/deit}}, with the default
training parameters, but without exponential moving averaging of the weights. More precisely, we trained for 300 epochs over 16 GPUs, with a batch size per GPU of 64, AdamW~\citep{loshchilov2017decoupled}, a weight decay of 0.05, learning rate of $5\mathrm{e}{-}4$, cosine
learning rate scheduler, a learning rate warmup from $1\mathrm{e}{-}6$ over 5 epochs and label smoothing~\citep{szegedy2016rethinking}. As data augmentation, we used color-jitter, random erase, and either cutmix or mixup~\citep{zhang2017mixup}.

For \diffq, we tested the penalty $\lambda$ in $\{1\mathrm{e}{-}3, 1\mathrm{e}{-2}, 0.1, 0.5, 1, 5\}$,
and group size $g$ in $\{1, 4, 8\}$. We use a minimum number of bits of 3, instead of 2, as this led to better stability. 
We use Adam~\citep{adam} to optimize the bits parameters, with a learning rate of $5\mathrm{e}{-}{4}$.

\noindent{\bf ImageNet - ResNet18 and ResNet50.}
We trained all models (\diffq and LSQ) for 400 epochs over 4 GPUs, with a batch size per GPU of 256, using  RMSProp~\cite{tieleman2012lecture} as implemented in the \texttt{timm} package\footnote{\url{https://github.com/rwightman/pytorch-image-models}} (also experimented with SGD however RMSProp provides better results), a weight decay of 0.05, learning rate of $5\mathrm{e}{-}4$, where we multiply the learning by 0.9875 after every epoch. We used a learning rate warmup from $1\mathrm{e}{-}6$ over 3 epochs and label smoothing~\citep{szegedy2016rethinking} with smoothing factor of 0.3. As data augmentation, we used color-jitter, random erase and cutmix using $\beta=0.2$ with probability of 0.3.

For \diffq, we tested the penalty $\lambda$ in $\{1\mathrm{e}{-}2, 3\mathrm{e}{-2}, 4\mathrm{e}{-2}, 5\mathrm{e}{-2}, 8\mathrm{e}{-2}, 0.1\}$, and group size $g$ in $\{8, 16\}$. We use a minimum number of bits of 2. We use Adam~\citep{adam} to optimize the bits parameters, with a learning rate of $5\mathrm{e}{-}{4}$.

\section{Supplementary results}
\label{app:results}

\subsection{ImageNet}
\label{supp:imagenet}
On Table~\ref{tab:imagenet_supp} results are reported for \diffq and LSQ~\cite{esser2020learned} using ResNet-18 and ResNet-50 on ImageNet dataset. We compared different model sizes and compression rates. Results suggest that \diffq is superior both in terms of accuracy and smaller model size. 

Results marked with * are the ones reported in \cite{esser2020learned} using slightly better uncompressed model. For fair comparison we reported both numbers. Notice, \diffq achieves comparable and even superior results over LSQ also under considering this setting.

\begin{table}[t!]
\begin{center}
\begin{small}
\caption{Additional comparison between \diffq and LSQ~\cite{esser2020learned} for different model sizes and compression rates. Results marked with * are the ones reported in \citet{esser2020learned} using slightly better uncompressed baseline model. For fair comparison we reported both numbers. All results reported are for the best model accuracy.}
\label{tab:imagenet_supp}
\centering
\begin{sc}
\resizebox{0.7\textwidth}{!}{
\begin{tabular}{c|c|c|c}
\toprule
Model       & Method     & Top-1 Acc. (\%) & M.S. (MB)   \\ \midrule
\multicolumn{4}{c}{ImageNet}                 \\ \midrule
ResNet-18   & Uncompressed                  & 70.9       & 44.6  \\ 
\midrule
ResNet-18   & LSQ 8 bits~\cite{esser2020learned}        & 71.8       & 11.2 \\  
ResNet-18   & LSQ 4 bits~\cite{esser2020learned}        & 70.7       & 5.6  \\ 
ResNet-18   & LSQ 3 bits~\cite{esser2020learned}        & 69.0         & 4.2  \\ 
\midrule
ResNet-18   & LSQ* 8 bits~\cite{esser2020learned}        & 71.1       & 11.2 \\  
ResNet-18   & LSQ* 4 bits~\cite{esser2020learned}        & 71.1       & 5.6  \\ 
ResNet-18   & LSQ* 3 bits~\cite{esser2020learned}        & 70.2         & 4.2  \\ 
\midrule
ResNet-18   & \diffq~(Ours)       & 71.8       & 7.6  \\ 
ResNet-18   & \diffq~(Ours)       & 71.1       & 5.3  \\ 
ResNet-18   & \diffq~(Ours)       & 70.2       & 4.5  \\ 
ResNet-18   & \diffq~(Ours)       & 69.7       & 4.1  \\ 
\midrule
ResNet-50   & Uncompressed                       & 77.1       & 97.5  \\ 
\midrule
ResNet-50   & LSQ 8 bits~\cite{esser2020learned}        & 76.8       & 24.5 \\ 
ResNet-50   & LSQ 4 bits~\cite{esser2020learned}        & 76.2       & 12.3 \\ 
ResNet-50   & LSQ 3 bits~\cite{esser2020learned}        & 75.6       & 9.3  \\ 
\midrule
ResNet-50   & LSQ* 8 bits~\cite{esser2020learned}        & 76.8       & 24.5 \\ 
ResNet-50   & LSQ* 4 bits~\cite{esser2020learned}        & 76.7       & 12.3 \\ 
ResNet-50   & LSQ* 3 bits~\cite{esser2020learned}        & 75.8       & 9.3  \\ 
\midrule
ResNet-50   & \diffq~(Ours)       & 76.9       & 14   \\ 
ResNet-50   & \diffq~(Ours)       & 76.6       & 10.5 \\ 
ResNet-50   & \diffq~(Ours)       & 76.3       & 8.8    \\ 
\bottomrule
\end{tabular}}
\end{sc}
\end{small}
\end{center}
\end{table}

\begin{table}[t!]
\caption{Detailed results of QAT and \diffq on the CIFAR-10/100 datasets. For each architecture and dataset,
we provide the performance of the baseline, QAT models with 2 to 4 bits, and two \diffq runs: v1. is the smallest model that is within a small range of the baseline performance, v2. is the best model of comparable size with QAT 2 bits, selected from the pool of candidates described in Section \ref{supp:image}. For Wide-ResNet, we report a single variant of \diffq, as it is both the smallest and the one with the best accuracy.}
\label{tab:cifar_supp}
\vskip 0.15in
\begin{center}
\begin{small}
\begin{sc}
\resizebox{0.9\textwidth}{!}{
\begin{tabular}{l|l|cc|cc|cc}
\toprule
	&	 & \multicolumn{2}{c}{MobileNet} & \multicolumn{2}{c}{ResNet-18} & \multicolumn{2}{c}{WideResNet} \\
\midrule
	&	 & Acc. (\%)~$\uparrow$ &  M. S. (MB)~$\downarrow$ & Acc. (\%)~$\uparrow$ &  M. S. (MB)~$\downarrow$ & Acc. (\%)~$\uparrow$ &  M. S. (MB)~$\downarrow$ \\
\midrule
\multirow{6}{*}{\rotatebox[origin=c]{90}{CIFAR-10}}
& Uncompressed    	        & 90.9 		   & 12.3 & 95.3  & 42.7  & 95.3    & 139.2\\
\cmidrule(lr){2-8}
& QAT 2bits    				& 78.1         & 0.88 & 87.2  & 2.70  & 70.8    & 8.81  \\
& QAT 3bits    				& 88.2         & 1.26 & 94.0  & 4.03  & 94.3    & 13.16 \\
& QAT 4bits    				& 90.1         & 1.64 & 95.0  & 5.36  & 94.4    & 17.50 \\
\cmidrule(lr){2-8}
& LSQ 2 bits                & 10.0         & 0.88 & 95.0  & 2.70  & 81.9. & 8.81  \\
& LSQ 3 bits                & 90.8         & 1.26 & 95.3  & 4.03  & 88.8  & 13.16 \\
& LSQ 4 bits                & 90.9         & 1.64 & 95.2  & 5.36  & 89.9 & 17.50 \\
\cmidrule(lr){2-8}
 & \diffq v1 & 90.3 & 0.94 & 94.9 & 3.17 & 94.1 & 8.81\\
 & \diffq v2 & 87.9 & 0.91 & 93.9 & 2.71 & 94.1 & 8.81\\
\midrule\midrule
\multirow{6}{*}{\rotatebox[origin=c]{90}{CIFAR-100}}
& Uncompressed    	            & 68.1 		   & 12.6 & 77.9 & 42.8 & 76.2  & 139.4\\
\cmidrule(lr){2-8}
& QAT 2bits    		            & 10.9         & 0.91  & 58.7 & 2.72  & 46.5  & 8.83  \\
& QAT 3bits    		            & 59.7         & 1.29  & 73.7 & 4.05  & 75.0  & 13.18 \\
& QAT 4bits    		            & 66.9         & 1.69  & 77.3 & 5.39  & 75.5  & 17.53 \\
\cmidrule(lr){2-8}
& LSQ 2 bits                    & 64.9         & 0.91 & 77.5  & 2.72  & 40.9  & 8.82  \\
& LSQ 3 bits                    & 67.7         & 1.29 & 77.7  & 4.05  & 55.6  & 13.18 \\
& LSQ 4 bits                    & 68.5         & 1.69 & 77.8  & 5.39  & 56.5  & 17.53 \\
\cmidrule(lr){2-8}
 & \diffq v1 & 68.5 & 1.10 & 77.6 & 4.82 & 75.3 & 8.83\\
 & \diffq v2 & 64.6 & 0.94 & 71.7 & 2.72 & 75.6 & 8.84\\
\bottomrule
\end{tabular}}
\end{sc}
\end{small}
\end{center}
\vskip -0.1in
\end{table}

\subsection{CIFAR-10/100}
\label{supp:cifar}
We report on Table~\ref{tab:cifar_supp} the results on the CIFAR10/100 datasets, which are shown for CIFAR100 in Figure~\ref{fig:cifar} in the main paper. Results are presented using MobileNet-v1, ResNet-18, and WideResNet. For CIFAR100 the presented results used for creating Figure~\ref{fig:cifar} in the main paper.
As we cannot show all the \diffq runs, we selected for each model and dataset two versions: v1 is the smallest model that has an accuracy
comparable to the baseline (accuracy is greater than $1 - 1/100$ times the baseline accuracy), while v2 is the model with the highest accuracy
that is comparable in size with the QAT 2 bits model (size must be smaller than $1 + 1/100$ times the baseline size, except for MobileNet, for which we had to allow a 4\% relative increase in size. The penalty and group size selected with this procedure is displayed on Table~\ref{table_hyper_supp}.

\begin{table}[t!]
\caption{Penalty $\lambda$ and group size $g$ for the v1 and v2 \diffq models reported on Table~\ref{tab:cifar_supp}}
\label{table_hyper_supp}
\vskip 0.15in
\begin{center}
\begin{small}
\begin{sc}
\resizebox{0.5\textwidth}{!}{
\begin{tabular}{l|l|cc|cc|cc}
\toprule
	&	 & \multicolumn{2}{c}{MobileNet} & \multicolumn{2}{c}{ResNet-18} & \multicolumn{2}{c}{WideResNet} \\
\midrule
	&	 & $\lambda$ & $g$ & $\lambda$ & $g$ & $\lambda$ & $g$ \\
\midrule
\multirow{2}{*}{CIFAR-10}
 & \diffq v1 & 1 & 16 & 0.1 & 8 & 5 & 16\\
 & \diffq v2 & 5 & 8 & 5 & 4 & 5 & 16\\
\midrule\midrule
\multirow{2}{*}{CIFAR-100}
 & \diffq v1 & 1 & 16 & 0.05 & 4 & 5 & 16\\
 & \diffq v2 & 5 & 16 & 5 & 8 & 1 & 16\\
\bottomrule
\end{tabular}}
\end{sc}
\end{small}
\end{center}
\vskip -0.1in
\end{table}

Looking first at v1 models, we achieve on all tasks and datasets
a model that is competitive with the baseline (sometimes even better),
with a model size that is smaller than a QAT 4 bits model (for instance more than 2MB saved on a ResNet-18 trained on CIFAR-10 compared to QAT 4 bits, for the same accuracy).

Now for v2, first note that as the minimum number of bits used by \diffq is exactly 2, it is not possible here to make a model smaller than QAT 2 bits. However, even with as little as 0.01 MB extra, \diffq can get up to 30\% increase in accuracy compared to QAT 2 bits (for a Wide ResNet). On all architectue and datasets,
the gain from \diffq over QAT 2 bits is at least 10\% accuracy.
This confirms in practice the bias of STE-based methods when the number
of bits is reduced, a bias that we already demonstrated in theory in Section~\ref{sec:counter}. In particular, it is interesting that the largest
improvement provided by \diffq is for the Wide ResNet model, which should be the easiest to quantize. But having the largest number of weights, it 
also likely the one that is the most sensitive to the oscillations of QAT quantized weights described in Section~\ref{sec:counter}.

\paragraph{Ablation.} Table~\ref{supp:tab:app_image_abb} summarizes the results of comparing QAT against \diffq for model quantization using a fixed number of bits using MobileNet, ResNet-18, and WideResNet on both CIFAR10 and CIFAR100. \diffq outperforms QAT, where this is especially noticeable while using 2 bits quantization, in which training is less stable for QAT. 

Next, we evaluated the affect of the group-size, $g$, on model size and accuracy, by optimizing \diffq models using $g \in \{1, 4, 8, \infty \}$. When $g=\infty$, we use a single group for the entire layer. Results for ResNet-18 using CIFAR-100 are summarized in Figure~\ref{fig:supp} (a). Interestingly, we observed that increasing $g$, yields in a smaller model size on the expanse of a minor decrease in performance. However, when setting $g=\infty$ model performance (model size and accuracy) is comparable to $g=8$ for this task.

\begin{table*}[t!]
\caption{A comparison between QAT and \diffq while we consider a fixed number of bits for all model parameters, specifically using 2, 3, and 4 bits. Results are reported for CIFAR-10 and CIFAR-100 using MobileNet-v1, ResNet-18. and WideResNet. We report Accuracy (Acc.) and Model Size (M.S.).}
\label{supp:tab:app_image_abb}
\vskip 0.15in
\begin{center}
\begin{small}
\begin{sc}
\resizebox{1.0\textwidth}{!}{
\begin{tabular}{l|l|cc|cc|cc}
\toprule
	&	 & \multicolumn{2}{c}{MobileNet} & \multicolumn{2}{c}{ResNet-18} & \multicolumn{2}{c}{WideResNet} \\
\midrule
	&	 & Acc. (\%)~$\uparrow$ &  M. S. (MB)~$\downarrow$ & Acc. (\%)~$\uparrow$ &  M. S. (MB)~$\downarrow$ & Acc. (\%)~$\uparrow$ &  M. S. (MB)~$\downarrow$ \\
\midrule
\multirow{9}{*}{\rotatebox[origin=c]{90}{CIFAR-10}}
& Uncompressed    	        & 90.9 		   & 12.3 & 95.3  & 42.8  & 95.3    & 139.4\\
\cmidrule(lr){2-8}
& QAT 2bits    				& 78.1         & 0.88 & 87.2  & 2.70  & 70.8    & 8.81  \\
& QAT 3bits    				& 88.2         & 1.26 & 94.0  & 4.03  & 94.3    & 13.16 \\
& QAT 4bits    				& 90.1         & 1.64 & 95.0  & 5.36  & 94.4    & 17.50 \\
\cmidrule(lr){2-8}
& LSQ 2 bits                & 10.0         & 0.88 & 95.0  & 2.70  & 81.9 & 8.81  \\
& LSQ 3 bits                & 90.8         & 1.26 & 95.3  & 4.03  & 88.8  & 13.16 \\
& LSQ 4 bits                & 90.9         & 1.64 & 95.2  & 5.36  & 89.9 & 17.50 \\
\cmidrule(lr){2-8}
& \diffq 2bits	 			& 84.1  	   & 0.88 & 92.3  & 2.70  & 94.4    & 8.81   \\
& \diffq 3bits				& 89.7         & 1.26 & 94.4  & 4.03  & 94.4    & 13.16  \\
& \diffq 4bits				& 90.4         & 1.64 & 95.1  & 5.36  & 94.6    & 17.50  \\
\midrule\midrule
\multirow{9}{*}{\rotatebox[origin=c]{90}{CIFAR-100}}
& Uncompressed    	            & 68.1 		   & 12.6 & 77.9 & 42.8 & 76.2  & 139.4\\
\cmidrule(lr){2-8}
& QAT 2bits    		            & 10.9         & 0.91  & 58.7 & 2.72  & 46.5  & 8.82  \\
& QAT 3bits    		            & 59.7         & 1.29  & 73.7 & 4.05  & 75.0  & 13.18 \\
& QAT 4bits    		            & 66.9         & 1.69  & 77.3 & 5.39  & 75.5  & 17.53 \\
\cmidrule(lr){2-8}
& LSQ 2 bits                    & 64.9         & 0.91 & 77.5  & 2.72  & 40.9  & 8.82  \\
& LSQ 3 bits                    & 67.7         & 1.29 & 77.7  & 4.05  & 55.6  & 13.18 \\
& LSQ 4 bits                    & 68.5         & 1.69 & 77.8  & 5.39  & 56.5  & 17.53 \\
\cmidrule(lr){2-8}
& \diffq 2bits	 			    & 17.2  	   & 0.91 & 66.6 & 2.72  & 72.8   & 8.82   \\
& \diffq 3bits				    & 60.1         & 1.29 & 76.7 & 4.05  & 76.9   & 13.18  \\
& \diffq 4bits				    & 66.8         & 1.69 & 77.5 & 5.39  & 76.9   & 17.53  \\
\bottomrule
\end{tabular}}
\end{sc}
\end{small}
\end{center}
\end{table*}

\subsection{EfficientNet-b3 on ImageNet}
\label{supp:imnet}
\begin{figure}[t!]
\centering 
\subfigure[]{\label{fig:gs}\includegraphics[width=0.45\columnwidth]{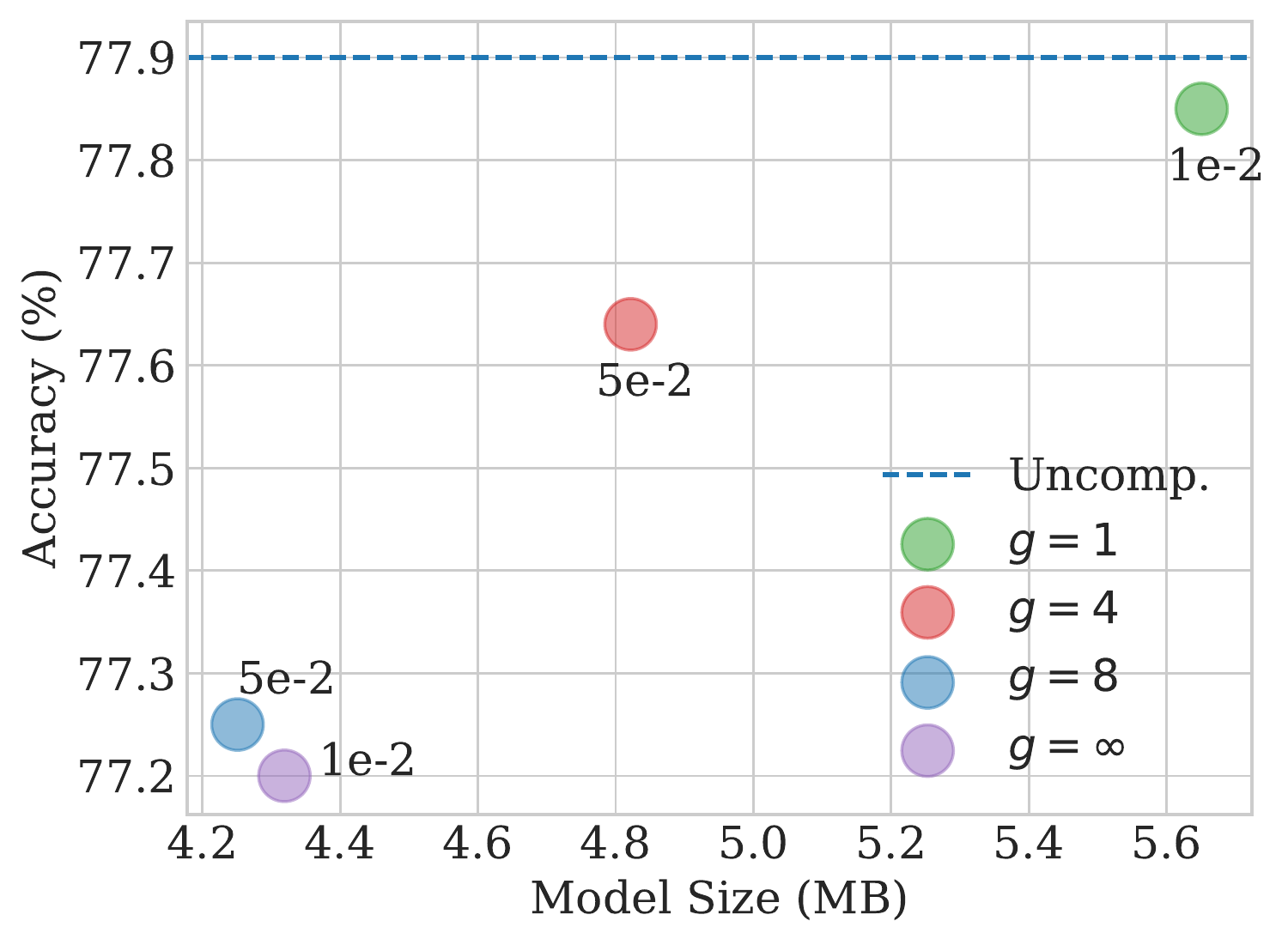}}
\subfigure[]{\label{fig:teaser}\includegraphics[width=0.48\columnwidth]{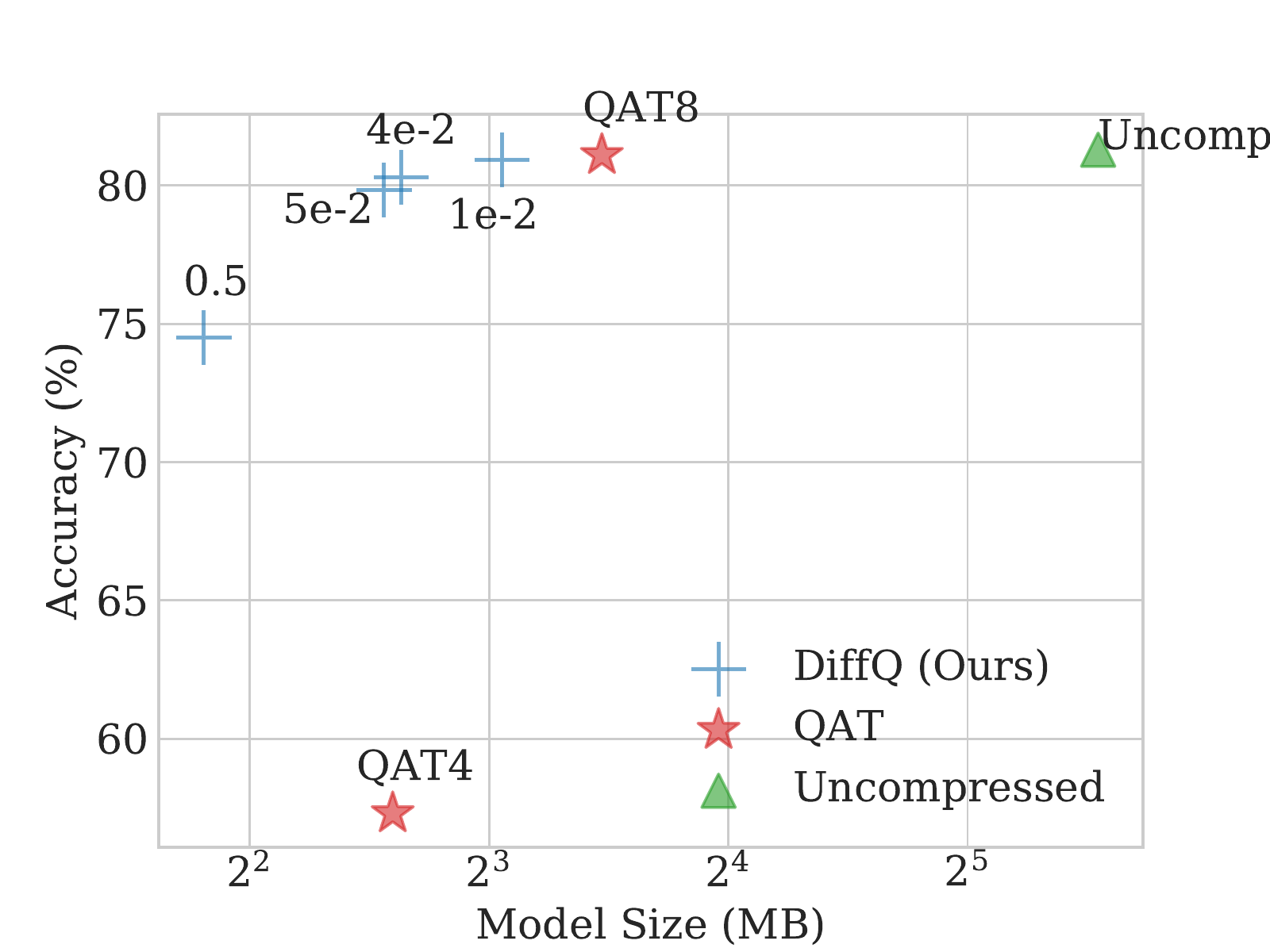}}
\caption{\textbf{(a)}: \diffq results with various groups sizes ($g \in \{1, 4, 8, \infty \}$). $g=\infty$ refers to a single group for the entire layer. For reference, we report the accuracy of the uncompressed model (42.8 MB). 
Models are Resnet-18 trained on CIFAR-100. 
\textbf{(b)}: ImageNet results using EfficientNet-B3 model. We plot the model size vs. model accuracy using different penalty levels. We additionally, present the uncompressed models (uncomp.) and Quantization Aware Training (QAT) using 4 and 8 bits.}
\label{fig:supp}
\end{figure}

On Table~\ref{supp:imagenet} we report the results for training
EfficientNet-b3~\cite{tan2019efficientnet} on the ImageNet dataset, matching the results reported on Figure~\ref{fig:teaser}.

\begin{table}[t!]
\caption{Image classification results for the ImageNet benchmark. Results are presented for \diffq and QAT using 4 and 8 bits using the EfficientNet-b3 model~\citep{tan2019efficientnet}. We report Top-1 Accuracy (Acc.) together with Model Size (M.S.).}
\label{supp:imagenet}
\vskip 0.15in
\begin{center}
\begin{small}
\begin{sc}
\begin{tabular}{l|cc}
\toprule
		& Top-1 Acc. (\%)~$\uparrow$ & M.S. (MB)~$\downarrow$ \\
\midrule
Uncompressed    	                & 81.6 		   & 46.7 \\
\midrule
QAT 4bits    		                &  57.3        & 6.3  \\
QAT 8bits    		                & 81.3         & 12.0 \\
PQ ~\citep{fan2020training} & 80.0 & \textbf{3.1} \\
\midrule
\diffq ($\lambda{=}0.05$)  	                & 80.8      & 6.0 \\
\diffq ($\lambda{=}0.01$)    	    & \textbf{81.5}      & 8.7  \\
\bottomrule
\end{tabular}
\end{sc}
\end{small}
\end{center}
\end{table}

As previously, we selected two versions of \diffq, one matching the size of QAT 8bits, and one smallest than QAT 4 bits. At 8 bits, \diffq achieves the same accuracy as the uncompressed baseline, for a slightly smaller model than QAT 8bits. As we lower the number of bits, we again see a clear advantage for \diffq, with both a smaller model (5.7MB against 6.1MB) than QAT 4bits, and significantly higher accuracy (76.8\% vs. 57.3\%).


The lower accuracy for QAT4 on ImageNet led us to take a closer look at the model performance. Figure~\ref{fig:supp} (b) depicts the model accuracy as a function of the number of epochs for both QAT4 and \diffq. Notice, similarly to the toy example presented in Section~\ref{sec:counter} training with QAT4 creates instability in the model optimization (especially near model convergence), which leads to significant differences in performance across adjacent epochs. When considering \diffq, model optimization is stable and no such differences are observed. 


    

\subsection{Activation Quantization for Language Modeling}

In Table~\ref{tab:app_text} we report language modeling results for a 16-layers Transformer models while applying activation quantization. Unlike the results in Table~\ref{tab:text} where we used per-channel activation quantization, here we report results with a histogram quantizer. Additionally when considering histogram quantizer, results suggest \diffq is superior to both QAT and QN when considering both model size and model performance.
\begin{table}[t!]
\caption{
Language modeling results for a 16 layer Transformer trained on Wikitext-103. We also test combining weight and activation quantization using a histogram quantizer. We compared \diffq to QAT and Quant-Noise (QN) method proposed by~\citet{fan2020training} (models with $\dagger$ were trained with layer drop of 0.2~\cite{fan2019reducing}, and 0.1 for the others.).}
\label{tab:app_text}
\vskip 0.15in
\begin{center}
\begin{small}
\begin{sc}
\resizebox{0.5\columnwidth}{!}{
\begin{tabular}{llcc}
\toprule
	Weights & Activation	 & PPL~$\downarrow$ & M. S. (MB)~$\downarrow$ \\
\midrule
Uncompressed (Ours) & -  	   & 18.1 & 942\\
QAT 8bits    & -	& 18.2           & 236 \\
QAT 4bits   & - 	&   28.8         & 118 \\
\diffq  ($\lambda{=}1, g{=}16$) & - & \textbf{18.0}          & 182 \\
\diffq  ($\lambda{=}10, g{=}16$) & - & 18.5          & \textbf{113} \\
\midrule
    8 bits & 8 bits    & 19.5  & 236\\
QAT 8bits & 8 bits   	& 26.0  & 236 \\
QAT 4bits & 8 bits    	&   34.6   & 118 \\
\diffq  ($\lambda{=}1, g{=}16$) & 8 bits & 19.1 & 182\\
\diffq  ($\lambda{=}10, g{=}16$) & 8 bits & 19.2 & \textbf{113}\\
\midrule
Uncompressed $\dagger$ & -  	      & 18.3 & 942\\
QN 8 bits$\dagger$    & QN 8 bits & \textbf{18.7} & 236 \\
QN 4 bits$\dagger$    & QN 8 bits & 20.5 & 118 \\ 
\bottomrule
\end{tabular}}
\end{sc}
\end{small}
\end{center}
\vskip -0.1in
\end{table}

\subsection{Uniform Noise vs. Gaussian Noise}
In Table~\ref{supp:noise} we provide an empirical comparison between uniform noise and Gaussian noise using ResNet-18 on CIFAR10 for \diffq. We found that using Gaussian noise acheives the same model size with better accuracy levels.

\begin{table}[t!]
\caption{Empirical comparison between uniform noise and Gaussian noise using ResNet-18 on CIFAR10 for \diffq.}
\label{supp:noise}
\vskip 0.15in
\begin{center}
\begin{small}
\begin{sc}
\begin{tabular}{l|cc}
\toprule
		Noise distribution & Top-1 Acc. (\%)~$\uparrow$ & M.S. (MB)~$\downarrow$ \\
\midrule
Uniform  	    & 86.9      & 2.7 \\
Gaussian  	    & 93.6      & 2.7  \\
\bottomrule
\end{tabular}
\end{sc}
\end{small}
\end{center}
\end{table}

\end{document}